\documentclass[12pt]{article}
\usepackage{format-and-macros/packages}
\usepackage{format-and-macros/statistics-macros}
\usepackage{format-and-macros/editing-macros}
\usepackage{format-and-macros/formatting}
\usepackage{fullpage}

\makeatletter
\let\@fnsymbol\@arabic
\makeatother

\title{
\vspace{-2em}SLOE: A Faster Method for Statistical Inference in High-Dimensional Logistic Regression}
\author{Steve Yadlowsky\thanks{Google Research, Brain Team}~,~Taedong Yun\thanks{Google Health}~,~Cory McLean\footnotemark[2]~,~Alexander D'Amour\footnotemark[1]\\
 \texttt{\normalsize\{yadlowsky,tedyun,cym,alexdamour\}@google.com}\vspace{-2em}}
\date{}

\begin{document}
\maketitle

\begin{abstract}
Logistic regression remains one of the most widely used tools in applied statistics, machine learning and data science. However, in moderately high-dimensional problems, where the number of features $d$ is a non-negligible fraction of the sample size $n$, the logistic regression maximum likelihood estimator (MLE), and statistical procedures based the large-sample approximation of its distribution, behave poorly. Recently, \citet{SurCa2019} showed that these issues can be corrected by applying a new approximation of the MLE's sampling distribution in this high-dimensional regime. Unfortunately, these corrections are difficult to implement in practice, because they require an estimate of the \emph{signal strength}, which is a function of the underlying parameters $\beta$ of the logistic regression. To address this issue, we propose SLOE, a fast and straightforward approach to estimate the signal strength in logistic regression. The key insight of SLOE is that the \citet{SurCa2019} correction can be reparameterized in terms of the \emph{corrupted signal strength}, which is only a function of the estimated parameters $\what \beta$. We propose an estimator for this quantity, prove that it is consistent in the relevant high-dimensional regime, and show that dimensionality correction using SLOE is accurate in finite samples. Compared to the existing ProbeFrontier heuristic, SLOE is conceptually simpler and orders of magnitude faster, making it suitable for routine use. We demonstrate the importance of routine dimensionality correction in the Heart Disease dataset from the UCI repository, and a genomics application using data from the UK Biobank. We provide an open source package for this method, available at \url{https://github.com/google-research/sloe-logistic}.
\end{abstract}

\section{Introduction}
\label{sec:intro}
Logistic regression is a workhorse in statistics, machine learning, data science, and many applied fields.
It is a generalized linear model that models a binary scalar outcome $Y \in \{0, 1\}$ conditional on observed features $X \in \mathbb R^d$ via
\begin{equation}
    \E[Y \mid X=x] = g(\beta^\top x),~\text{with}~g(t) := \frac{1}{1 + \exp(-t)},\label{eq:logistic regression}
\end{equation}
with the coefficients $\beta$ fit using observed data.
Logistic regression is popular as a scientific tool because the model is often accurate, and comes with well-established statistical inference procedures for quantifying uncertainty about the parameters $\beta$ and predictions $g(\beta^\top x)$ at test inputs $x$.
For example, most statistical software packages not only produce predictions from the model, but also summaries such as confidence intervals (CIs) and $p$-values that enable practitioners to understand the strength of evidence for the prediction in a quantitative way.
These widely adopted estimation and statistical inference routines are based on approximations justified by large-sample asymptotic theory of the maximum likelihood estimator (MLE) $\what \beta$.
These approximations are derived from the limiting distribution of $\what \beta$ as the sample size $n$ tends toward infinity, but the number of covariates $d$ remains fixed.

\begin{figure*}[t]
	\centering
	\begin{subfigure}[t]{0.47\columnwidth}
		\centering
    \includegraphics[width=\columnwidth]{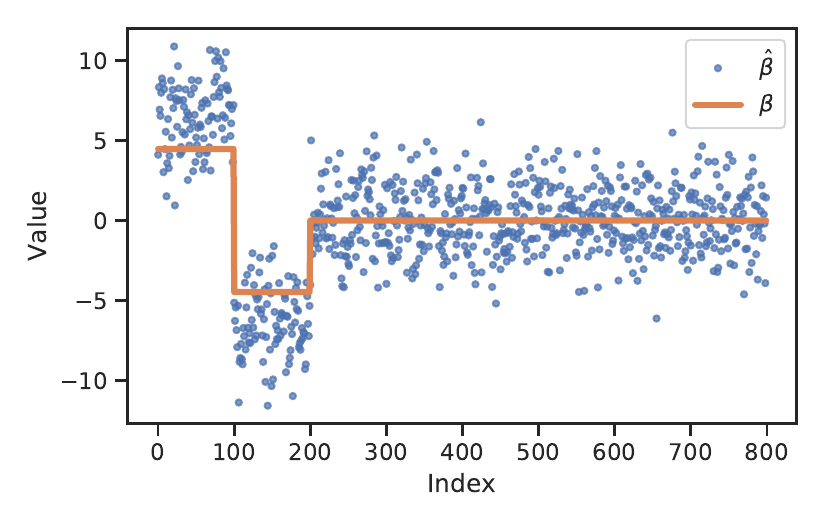}%
    \caption{True and estimated coefficients}
    \label{fig:simulation-coef}	
	\end{subfigure}
	\begin{subfigure}[t]{0.47\columnwidth}
		\centering
    \includegraphics[width=\columnwidth]{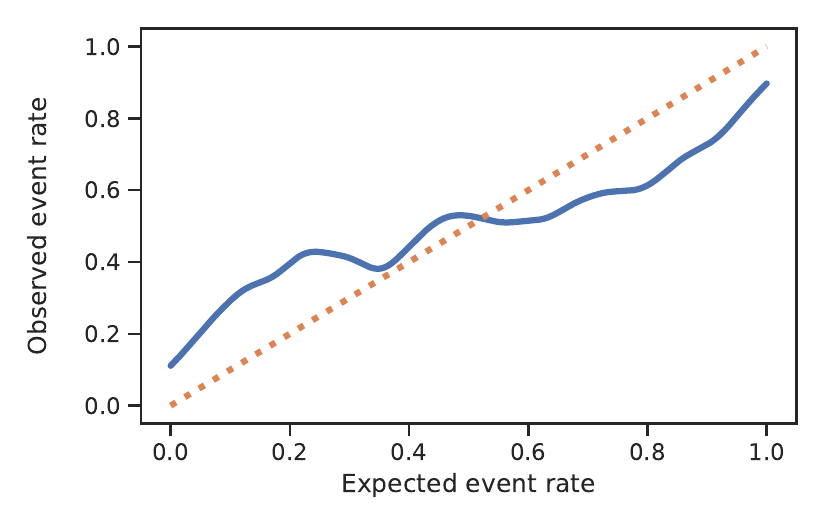}
    \caption{Calibration curve for the model predictions on an iid test set.}
    \label{fig:calib-mle}
	\end{subfigure}
	\caption{Coefficients and calibration of the predictions from the MLE from simulated data with $n=4000,~d = 800$. Neither the coefficients, nor predictions are converging to their true values as predicted by standard large-sample asymptotic theory.}\label{fig:model-snapshot}
	\vspace{-0.15in}
\end{figure*}

Unfortunately, these standard approximations perform poorly when the number of covariates $d$ is of a similar order of magnitude to the number of samples $n$, even when the sample size is large \citep{FanDeLv19}.
Recently, \citet{SurCa2019} showed that, in this setting, the behavior of $\what \beta$ in finite samples is better approximated by its limiting distribution in another asymptotic regime, where the aspect ratio $\kappa := d / n > 0$ is held fixed as both $n$ and $d$ grow to infinity.
Here, they show that the estimated coefficients $\what \beta$ (and therefore, the predictions) have systematically inflated magnitude and larger variance than the standard approximation predicts.
The precise characterization of the limiting distribution of $\what \beta$ in \citet{SurCa2019} and \citet{ZhaoSuCa2020} justify a new approximation, which, in principle, facilitates debiasing estimates and constructing CIs and $p$-values for parameters and test predictions, alike.

The goal of this paper is to make the application of these groundbreaking results practical.
A major barrier to adoption currently is that
the calculation of
bias and variance corrections in this framework requires knowing the signal strength $\gamma^2 := \var(\beta^\top X)$, which is challenging to estimate because it is a function of the unknown parameter vector $\beta$.
\citet{SurCa2019} proposed a heuristic called ProbeFrontier to estimate this quantity, but this approach is computationally expensive, conceptually complex, and hard to analyze statistically.
We propose a simpler estimation procedure.
Our approach reparameterizes the problem in terms of the corrupted signal strength parameter $\eta^2 := \lim_{n\to\infty} \var(\what \beta^\top X)$ that includes the noise in the estimate $\what{\beta}$. This is more straightforward (though non-trivial) to estimate.
We propose the Signal Strength Leave-One-Out Estimator (SLOE), which consistently estimates $\eta^2$, and show that using this to perform inference yields more accurate CIs in finite samples. 
Importantly, SLOE takes orders of magnitude less computation than ProbeFrontier, having similar runtime to the standard logistic regression fitting routine.

\section{Preliminaries}
\label{sec:prelim}
In this section, we revisit some fundamentals of statistical inference with logistic regression, review recent advances in high dimensional settings by \citet{SurCa2019} and \citet{ZhaoSuCa2020}, and discuss the implications of this paradigm shift in terms of better characterizing practice.

\subsection{Logistic Regression and Statistical Inference}

Estimates of $\beta := (\beta_1, \cdots, \beta_d)$ in the logistic regression model are usually obtained through maximum likelihood estimation,
by maximizing the empirical log-likelihood
\begin{align}
    \what \beta &:= \argmax_{\beta \in \R^d} \frac{1}{n} \sum_{i=1}^n Y_i \log(g(\beta^\top X_i)) + (1 - Y_i) \log(1-g(\beta^\top X_i)). 
\end{align}
The log likelihood is concave, and has a unique maximizer whenever the outcomes are not linearly separable in the covariates.
We will use logistic regression synonymously with
maximum likelihood estimation in the logistic regression model, and call $\what \beta$ the MLE.


The large-sample asymptotic statistical theory \citep{LehmannRo05} taught in nearly every university Statistics program characterizes the behavior of the estimated coefficients and the predictions made with them in the limit as $n \to \infty$ while holding the number of features $d$ fixed. Under this theory, estimates converge to their true value, $\what{\beta} \cp \beta$, and the estimation error $\what{\beta} - \beta$ and prediction error $g(\what{\beta}^\top x) - g(\beta^\top x)$ will be small to observe, unless amplified by a factor of $\sqrt{n}$, in which case $\sqrt{n}(\what{\beta} - \beta) \cd \normal{}(0, \mathcal I_{\beta}^{-1})$, where $\mathcal I_\beta := E[D_\beta XX^\top]$ is the Fisher information matrix, with $D_\beta:=g(\beta^\top X)(1-g(\beta^\top X))$.

Of course, when analyzing real data, data scientists only have access to a finite number of samples, and so this theory serves as an approximate characterization of the behavior expected in practice. If the approximation is good, one can make inferences about the underlying data generating distribution. For example, for $\delta \in (0,1)$, we can construct confidence intervals (CIs) that will contain the true parameters with probability $1 - \delta$. Propagating the uncertainty to predictions gives CIs for the outcome probabilities that can help contextualize machine learning predictions for users such as scientists or clinicians \citep{KompaSnBe21}.

The large sample approximation is a good one when the number of predictors is small relative to the number of observations. However, in modern scientific studies and data-driven technology, while many samples may be collected, many features of each sample are also collected. In such settings, the number of features $d$ may be significant compared to the sample size $n$. In such cases, the large sample characterization can be quite misleading.

We can visualize the extent to which this breakdown in assumptions leads to bias. Figure~\ref{fig:model-snapshot} shows the estimated coefficients and calibration curve for a logistic regression model fit on simulated data with $n = 4000$ and $p = 800$. Despite $n$ being reasonably large, these illustrate significant departures from the behavior predicted by the large sample asymptotics. The magnitude of the estimated coefficients is systematically overestimated and they are noisy. Additionally, the model is overconfident in its predictions. Despite this, the classification accuracy of the resulting model on the in-distribution test data is $0.70$, which is close to the optimal accuracy of $0.77$.

Recently, \citet{SurCa2019} and \citet{ZhaoSuCa2020} showed that the behavior of the MLE in high-dimensional settings is better characterized by a different asymptotic approximation.
In this approximation, the aspect ratio $\kappa = d / n$, instead of the number of predictors $d$, remains fixed as $n$ grows to infinity.
Formally, they study a sequence of problems in which:
\begin{itemize}
\vspace{-0.5em}
    \item $n,d \to \infty$; $d / n \to \kappa > 0$.
    \item $\gamma^2 = \var(\beta^\top X)$ is fixed as $d$ grows.
    \footnote{This prevents the expectations $g(\beta^\top X)$ from converging to 0 or 1 as $n$ grows. To satisfy this, the effect of each covariate shrinks as $n$ grows. It is complementary to sparsity models, being more realistic for many problems where predictive performance comes from pooling many weak signals.}
    \item The MLE exists asymptotically for $(\kappa, \gamma)$.
\vspace{-0.5em}
\end{itemize}
The semantics of this asymptotic regime are a bit more complicated than the classical one; we can no longer think of the argument in terms of a fixed problem with increasing data.
However, the characterization from this asymptotic theory better reflects the finite-sample behavior of the MLE $\what \beta$ when the data has a moderate aspect ratio $\kappa$.

For simplicity, we state the key result from this work in the case where the covariates are drawn from a spherical Gaussian, i.e., $X \sim \normal(0, I_d)$, but note that these results are generalized in \citet{ZhaoSuCa2020} to arbitrary covariance structures. So far, proof techniques only work for normally distributed features, however empirical results (Fig.~\ref{fig:coverage-gwas} and previous work) suggest a decent approximation for sub-Gaussian features.
In this regime, the distribution of the MLE $\what \beta$ converges to a normal distribution asymptotically centered around the inflated parameter $\alpha\beta$, for some $\alpha > 1$.
In particular, for any coefficient $\beta_j$,
\begin{align}
    \sqrt{n} \left(\what \beta_j - \alpha\beta_j\right) &\cd \normal{}(0, \sigma_\star^2),\label{eq:coefficient convergence}
\end{align}
and for the predicted logit of a test input\footnote{Note here that $x$ is a sequence in $n$. Up to a $1/\sqrt{d}$ scale factor, these constraints are satisfied for a fixed sequences of examples drawn from an iid test dataset with high probability.} $x \in \mathbb R^d$ with asymptotically finite norm and $\sqrt{n}x^\top \beta = O(1)$,
\begin{align}
    \sqrt{n}\|x\|_2^{-1} \left(\what \beta^\top x - \alpha\beta^\top x\right) &\cd \normal{}(0, \sigma_\star^2),\label{eq:prediction convergence}
\end{align}
for constants $\alpha$ and $\sigma_\star$, to be discussed later, that are only functions of $\kappa$ and $\gamma$.
The general convergence result, in which $X$ can have arbitrary covariance, is stated succinctly in Theorem 3.1 of \citet{ZhaoSuCa2020}.

\begin{figure}[t]
	\centering
		\includegraphics[width=0.6\columnwidth]{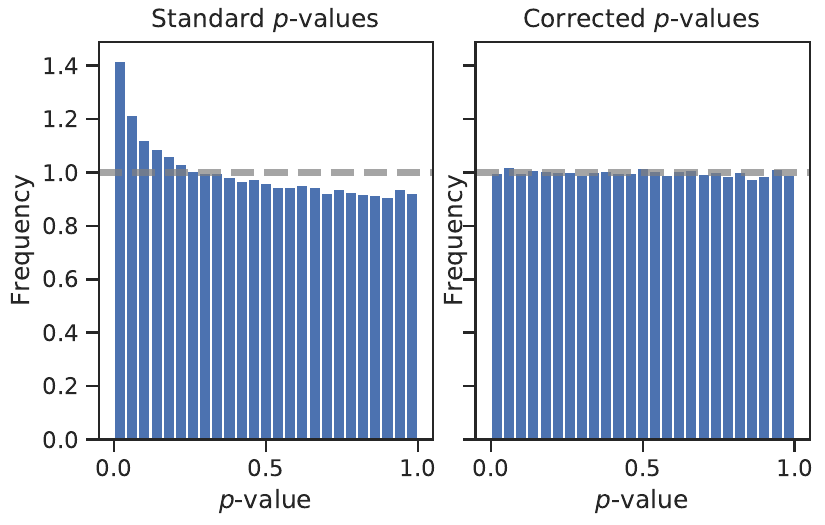}
		\caption{Histogram for $p$-values according to the standard and correct Wald test for the null coefficients in simulated data with $\kappa=0.1,~\gamma^2 = 5,~n=4000$.  }\label{fig:pvalues}
	\vspace{-0.15in}
\end{figure}

Based on this theory, if $\alpha$ and $\sigma_\star$ were known, one could construct corrected approximate CIs and $p$-values. For example, taking $\Phi^{-1}$ to be the quantile function of a standard normal distribution:
    \begin{gather}
        (1-\delta)~\text{CI for}~\beta_j:~\left(\frac{\what \beta_j}{\alpha} \pm \Phi^{-1}(1-\delta/2) \frac{\sigma_\star}{\alpha n^{1/2}}\right)
        \label{eq:ci-const}
        \\
        p\text{-value under }\beta_j = 0:~
        p = \Phi^{-1}(Z)~\text{where}~Z = \what \beta_j / \frac{\sigma_\star}{\alpha n^{1/2}}.
    \label{eq:pval-const}
    \end{gather}
In Figure~\ref{fig:pvalues}, we show that the $p$-values obtained based on these adjustments have the expected behavior (i.e., they are uniform when the null is true), whereas those obtained from the standard approximation give unexpected behavior.

\subsection{Calculating Dimensionality Corrections}
\citet{SurCa2019} show that the constants $\alpha$ and $\sigma_\star$ are determined by the solution to a system of three non-linear equations, which depend implicitly on the aspect ratio $\kappa$ and the signal strength $\gamma^2 := \var(\beta^\top X)$:
\begin{equation}
    \label{eq:nonlin-eq}
    \begin{cases}
    \kappa^2 \sigma_\star^2 &= \E\left[ 2 g(Q_1)(\lambda g(\prox_{\lambda G}(Q_2)))^2 \right] ,\\
    0 &= \E\left[ g(Q_1)Q_1 \lambda g(\prox_{\lambda G}(Q_2)) \right], \\
    1-\kappa &= \E\left[ \frac{2 g(Q_1)}{1 + \lambda g'(\prox_{\lambda G}(Q_2))} \right],
    \end{cases}
\end{equation}
where $\prox_{\lambda G}(s) = \argmin_{t} \lambda G(t) + \frac{1}{2}(s - t)^2$, for $G$ the anti-derivative of $g$, and
\begin{equation}
\label{eq:nonlin-dist}
   \left(\begin{array}{c}Q_1 \\ Q_2 \end{array}\right) \distas \normal\left(0, \left[ \begin{array}{cc}
    \gamma^2 & -\alpha \gamma^2 \\
    -\alpha \gamma^2 & \alpha^2 \gamma^2 + \kappa \sigma_\star^2
\end{array} \right] \right). 
\end{equation}
Here, $Q_1$ is a random variable with the same distribution (asymptotically) as the population logits $\beta^\top X$, and $Q_2$ is a random variable with the same distribution (asymptotically) as the logits with the biased MLE plugged in $\what \beta^\top X$.
The auxiliary parameter $\lambda$ corresponds to the limiting average eigenvalue of the inverse Hessian at the MLE, $(1/n \sum_{i=1}^n X_i g'(\what{\beta}^\top X_i) X_i^\top)^{-1}$, which is useful for making comparisons to asymptotic approximations from standard theory.
\citet{SurCa2019} note that this system has a unique solution so long as $(\kappa, \gamma)$ take values at which the MLE exists asymptotically.
Once set up, these equations are straightforward to solve with numerical methods.

The key quantity needed to set up these equations is the signal strength parameter $\gamma$, which is not directly observable.
Thus, applying dimensionality-corrected procedures in practice requires that $\gamma$ be estimated from data.
This is difficult, as $\gamma$ is itself a function of the unknown parameter vector $\beta$, and is the main focus of this paper.

\citet{SurCa2019} suggest a heuristic method, called ProbeFrontier, for estimating $\gamma$ in practice.
The idea is to search for the (sub)sample size $n' < n$ such that the observations in the subsample are linearly separable.
Sub-sampling the data changes the aspect ratio $\kappa$ without changing the signal strength $\gamma$. For a fixed signal strength $\gamma$, there is a sharp aspect ratio cutoff $\kappa_\star(\gamma)$ (the ``frontier'') above which the data is separable with high probability.
Based on these ideas, ProbeFrontier then inverts $\kappa_\star(\gamma)$ to estimate $\gamma$ from the empirical aspect ratio $p/n'$.
This requires subsampling the data repeatedly at various candidate aspect ratios, and for each subsample, checking whether the data are linearly separable using a linear program.
Repeatedly checking the separability is computationally expensive, and the statistical behavior near the frontier makes analysis tricky.

\section{A Practical Method for Dimensionality-Corrected Inference}

As we established above,
the key to estimating confidence intervals for $\beta$ or the predictions $g(x^\top \beta)$ is obtain a consistent estimate of $\gamma^2$ to plug in to Eqs.~\eqref{eq:nonlin-eq} and~\eqref{eq:nonlin-dist} to get estimates of $\alpha$ and $\sigma_\star$.
Here, we describe our method for obtaining such an estimate. First, we reparameterize the covariance in \eqref{eq:nonlin-dist} in terms of a more easily-estimated ``corrupted signal strength'' parameter, and then, we derive a computationally efficient estimator of this parameter by approximating a consistent leave-one-out estimator.

We implemented this estimator in Python, using scikit-learn to perform the MLE, and numpy / scipy \citep{Numpy20,Scipy20} for the high dimensional adjustment and inference. You can find our implementation, along with the simulation experiments described in Section~\ref{sec:simulation}, at \url{https://github.com/google-research/sloe-logistic}.

\subsection{Reparameterization}
\label{sec:reparam}
We begin by reparameterizing the estimating equations~\eqref{eq:nonlin-eq} and~\eqref{eq:nonlin-dist} in terms of a quantity that is easier to estimate.
As it appears in \citet{SurCa2019}, \eqref{eq:nonlin-dist} is written in terms of the signal strength $\gamma^2 = \lim_{n \to \infty} \var(\beta^\top X)$, but it can also be written in terms of $\eta^2 := \lim_{n\to\infty} \var(\hat \beta^\top X)$.
Proposition~2.1, and Lemmas~2.1,~3.1 from \citet{ZhaoSuCa2020} imply that $\eta^2 = \alpha^2 \gamma^2 + \kappa \sigma_\star^2$.
Thus, we can write \eqref{eq:nonlin-dist} equivalently as
\begin{equation}
\label{eq:nonlin-dist-eta}
    \left(\begin{array}{c}Q_1 \\ Q_2 \end{array}\right) \distas \normal\left(0, \left[ \begin{array}{cc}
    \eta - \kappa \sigma^2 & -\alpha \left(\eta^2 - \kappa \sigma_\star^2\right) \\
    -\alpha \left(\eta^2 - \kappa \sigma_\star^2\right) & \eta^2
\end{array} \right] \right).
\end{equation}
We can think of $\eta^2$ as the \emph{corrupted signal strength} of the predictors, corrupted by the fact that we only have an estimate $\what{\beta}$ of the true signal $\beta$ in our predictors. Note that $\var(\what{\beta}^\top X) = \E[\what{\beta}^\top \Sigma \what{\beta}]$, which suggests a natural strategy for estimate $\eta^2$ if $\Sigma$ were known. Shortly, we will define the (corrupted) Signal Strength Leave-One-Out Estimator (SLOE), $\what{\eta}_{\text{SLOE}}^2$, that estimates $\eta^2$ without knowing $\Sigma$.

By plugging this estimate from the data along with the aspect ratio $\kappa$ in to Eqs.~\eqref{eq:nonlin-eq} and~\eqref{eq:nonlin-dist-eta}, we can derive estimates $\what{\alpha},$ $\what{\sigma_\star^2},$ and $\what{\lambda}$ of the bias $\alpha$, variance $\sigma_\star^2$ and hessian inflation $\lambda$, respectively. Then, we can construct CIs \eqref{eq:ci-const} and compute $p$-values \eqref{eq:pval-const} using these estimates.

\subsection{SLOE Signal Strength Estimator}
\label{sec:SLOE}
We now describe how we estimate the corrupted signal strength $\eta^2 =
 \lim_{n\to\infty} \var(\hat \beta^\top X) = \lim_{n\to\infty} \E[\what \beta^\top \Sigma \what \beta]$ from data, using a computationally efficient approximation to an ideal, but impractical leave-one-out estimator.

The key challenge in estimating $\eta$ is that the predictor covariance $\Sigma$ must also be estimated. 
When $\what{\Sigma}$ and $\what{\beta}$ are estimated from the same set of observed $(X_i)_{i=1}^n$, these quantities have non-trivial dependence even in large samples, rendering the na\"ive estimate $\what \beta^\top \what \Sigma \what \beta$ asymptotically biased.

This dependence can be mitigated by a leave-one-out (LOO) estimator.
In particular, let $(\what{\beta}_{-i})_{i=1}^n$ be the set of $n$ LOO MLEs, each calculated with all of the data except the $i$-th example.
Define the LOO estimator as the empirical variance of the LOO logits $(\what \beta_{-i}^\top X_i)_{i=1}^n$:
\begin{equation}
    \what \eta_{\text{LOO}}^2 = \frac{1}{n} \sum_{i=1}^n (\what{\beta}_{-i}^\top X_i)^2 - \left(\frac{1}{n} \sum_{i=1}^n \what{\beta}_{-i}^\top X_i\right)^2. \label{eq:emp-var}
\end{equation}
This estimator eliminates the problematic dependence between $\what \beta$ and $\what \Sigma$ because $\what \beta_{-i}$ is independent of $X_i$.
Theorem~\ref{thm:consistency} shows that $\what \eta_{\text{LOO}}$ is consistent for $\eta$ (see the Supplementary Materials for proof).

\begin{theorem}
\label{thm:consistency}
Assume that $\gamma$ and $\kappa$ are in the range where the MLE exists asymptotically (see \citet[Theorem 1]{SurCa2019}), and that for each $n$, $p(n)$ satisfies $\lim_{n\to\infty} p / n = \kappa$. Let $\Sigma_p$ be a $p \times p$ positive definite matrix with a bounded condition number, for all $p$, and $\beta$ be a $p$-dimensional vector satisfying $\lim_{n\to\infty} \beta^\top \Sigma_p \beta = \gamma^2$, $\sqrt{n}\beta \cd \Pi$, a distribution with a second moment and $\text{Ave}_j n\beta_j^2 \to \E_\Pi[\beta^2]$. Assume that $X_i \simiid \normal{}(0, \Sigma_p)$, and $Y_i$ is drawn from the logistic regression model with parameter $\beta$ given $X_i$, independent of the other observations. 
Then, $\what{\eta}_{\text{LOO}}^2 \cp \eta^2$.
\end{theorem}

Unfortunately, this LOO estimator is impractical, as it requires refitting the logistic regression $n$ times. 
To address this, we define the \textbf{Signal Strength Leave-One-Out Estimator (SLOE)}.
\footnote{In the spirit of the method we leave one `O' out.}
SLOE replaces each LOO logit $\what \beta_{-i}^\top X$ with an approximation $S_i$, which can be expressed as a simple update to the full-data logit $\beta^\top X$:
\begin{gather}
    U_i = X_i^\top H^{-1} X_i \nonumber \\
    S_i = \what{\beta}_{}^\top X_i + \frac{U_i}{1 + g'(\what{\beta}^\top X_i) U_i}(Y_i - g(\what{\beta}^\top X_i)), \nonumber \\
    \what{\eta}_{\text{SLOE}}^2 = \frac{1}{n}\sum_{i=1}^n S_i^2 - \left(\frac{1}{n}\sum_{i=1}^n S_i\right)^2.
\end{gather}
To derive this estimator, we use a technique inspired by the proof for the asymptotic statistical properties of the MLE using leave-one-out estimators \citep{SurCa2019}. First, we write out the optimality condition for $\what{\beta}$ and $\what{\beta}_{-i}$ respectively. Let $\mathcal{I} = \{1, \dots, n\}$ be the indices of all the examples and $\mathcal{I}_{-i} = \{1,\dots,i-1, i+1, \dots, n\}$ be the indices of all but the $i$-th example. Then, by the first order optimality conditions for the MLE,
\begin{align*}
    & \sum_{j \in \mathcal{I}} X_j (Y_j - g(\what{\beta}^\top X_j)) = 0,~\mbox{and}~\sum_{j \in \mathcal{I}_{-i}} X_j(Y_j - g(\what{\beta}_{-i}^\top X_j)) = 0.
\end{align*}
Taking the difference between these two equations gives
\begin{align}
    0 &= X_i(Y_i - g(\what{\beta}^\top X_i))+ \sum_{j \in \mathcal{I}_{-i}} X_j(g(\what{\beta}_{-i}^\top X_j) - g(\what{\beta}^\top X_j)).  \label{eq:diff-of-est-eqs}
\end{align}
We expect the difference between $\what{\beta}_{-i}$ and $\what{\beta}$ to be small, so we can approximate the difference $g(\what{\beta}_{-i}^\top X_j) - g(\what{\beta}^\top X_j)$ well with a Taylor expansion of $g$ around $\what{\beta}^\top X_j$,
\begin{align}
    0 &\approx X_i(Y_i - g(\what{\beta}^\top X_i)) + \sum_{j \in \mathcal{I}_{-i}} X_j g'(\what{\beta}^\top X_j) X_j^\top (\what{\beta}_{-i}-\what{\beta}). \label{eq:taylor}
\end{align}

\begin{figure}[t]
\begin{center}
\centerline{\includegraphics[width=0.6\columnwidth]{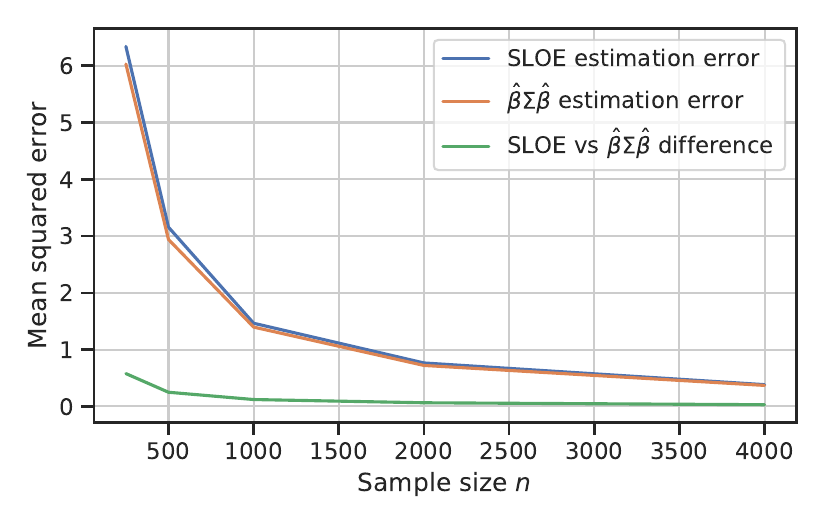}}
\caption{Empirical convergence of SLOE over sample sizes. Shows average squared differences over $1000$ simulations of $(\what{\eta}_{\SLOE}^2 - \eta^2)^2$, $(\what{\beta}^{\top}\Sigma \what{\beta} - \eta^2)^2$, and $(\what{\eta}_{\SLOE}^2 - \what{\beta}^{\top}\Sigma \what{\beta})^2$. The difference between $\what{\beta}^\top \Sigma \what{\beta} = \var(\what{\beta}^\top X)$ and its limit $\eta^2$ appears to be the dominant term.}
\vspace{-1em}
\label{fig:convergence}
\end{center}
\end{figure}

When the MLE exists, the empirical Hessian of the log likelihood for the full data and leave-one-out data, 
\begin{align*}
&H = -\sum_{j \in \mathcal{I}} X_j g'(\what{\beta}^\top X_j) X_j^\top~\mbox{and}~H_{-i} = -\sum_{j \in \mathcal{I}_{-i}} X_j g'(\what{\beta}^\top X_j) X_j^\top,
\end{align*}
respectively, are positive definite. Therefore, we can solve for $\what{\beta}_{-i}-\what{\beta}$ to get
\begin{align*}
     \what{\beta}_{-i}-\what{\beta} \approx H_{-i}^{-1} X_i(Y_i - g(\what{\beta}^\top X_i)).
\end{align*}
Then, we can accurately approximate the term $\what{\beta}_{-i}^\top X_i$ as
\begin{equation*}
    \what{\beta}_{}^\top X_i + X_i^\top H_{-i}^{-1} X_i(Y_i - g(\what{\beta}^\top X_i)).
\end{equation*}
To estimate all of the matrix inverses efficiently, we can take advantage of the fact that they are each a rank one update away from the full Hessian,
\begin{equation*}
     H_{-i} = H + X_i g'(\what{\beta}^\top X_i) X_i.
\end{equation*}
Applying the Sherman-Morrison inverse formula \citep{ShermanMo50} gives
\begin{equation*}
    X_i^\top H_{-i}^{-1} X_i = \frac{X_i^\top H^{-1} X_i}{1 + g'(\what{\beta}^\top X_i) X_i^\top H^{-1} X_i}.
\end{equation*}
Therefore, inverting one matrix gives us what we need from all $n$ inverse Hessians. Approximating $\what{\beta}_{-i}^\top X_i$ in $\what{\eta}^2_{\text{LOO}}$ with this gives us the SLOE estimator, $\what{\eta}^2_{\SLOE}$. To show that the SLOE estimator $\what{\eta}^2_{\SLOE}$ is consistent, we show the remainders from the Taylor approximation used to derive SLOE, \eqref{eq:taylor}, vanish in the limit (see the Supplement for proof).

\begin{proposition}
\label{prop:SLOE-loo}
Under the conditions of Theorem~\ref{thm:consistency}, the estimators $\what{\eta}_{\text{LOO}}$ and $\what{\eta}_{\SLOE{}}$ are asymptotically equivalent,
$\what{\eta}_{\text{LOO}} = \what{\eta}_{\SLOE{}} + o_P(1)$, and therefore, $\what{\eta}_{\SLOE{}} \cp \eta$.
\end{proposition}

\begin{remark}
The majority of the proof of Theorem~\ref{thm:consistency} allows us to derive a rate of convergence, $\what{\eta}^2 - \what{\beta}^\top \Sigma \what{\beta} = o_P(n^{1/4 - \epsilon})$, for any $\epsilon > 0$. However, to show that $\what{\beta}^\top \Sigma \what{\beta} \to \eta^2$, our proof relies on the approximate message passing construction of \citet{SurCa2019} and \citet{ZhaoSuCa2020}. Apparently, deriving rates of convergence for these estimators would require significant new technical results on the analysis of approximate message passing.
In Figure~\ref{fig:convergence}, we show in simulation that the estimation error in $\what \eta_{\SLOE}$ appears to be dominated by this latter difference, but that it appears to converge at the standard $n^{-1/2}$ rate.
\end{remark}

\section{Simulations}
\label{sec:simulation}
In this section, we evaluate $\what \eta_{\SLOE}$ in simulated data, where ground truth parameter values are known.
We show (1) that the asymptotic properties of $\what \eta_{\SLOE}$ hold in realistic sample sizes; (2) that $\what \eta_{\SLOE}$ can be used to make effective dimensionality corrections even when the covariates $X$ do not have a Gaussian distribution, as assumed in the theory; and (3) that $\what \eta_{\SLOE}$ requires orders of magnitude less computation than the current heuristic ProbeFrontier.

Here, we evaluate approximations in terms of the coverage rate of 90\% CIs for the prediction probabilities $\mu := g(\beta^\top x) = P(Y = 1 \mid X = x)$. 
In each simulation, we calculate a $90\%$ CI for $\mu_i$ for each observation $i$ in the corresponding test set, using the standard CI and the corrected CI, and compute the fraction of CIs that contain the true $\mu_i$.
Valid and efficient 90\% CIs will cover the target quantity 90\% of the time.
This evaluation simultaneously checks whether the CIs centering and width are appropriately calibrated, and whether the normal approximation adequately captures the tail behavior of the estimator.



In our simulations, we use a data generating process parameterized by the sample size $n$, the aspect ratio $\kappa$, and signal strength $\gamma^2$.
We draw $n$ examples of covariate vectors $X$ from a given distribution, and, conditional on $x_i$, calculate the $\mu_i$ according to the logistic regression model where $\beta$ is parameterized by $\gamma$ as follows,
\begin{equation*}
    \beta_j = \begin{cases}
    2 \gamma / \sqrt{d} & j \le d / 8 \\
    -2 \gamma / \sqrt{d} & d/8 < j \le d / 4\\
    0 & \text{otherwise,}
    \end{cases}
\end{equation*}
and finally, draw $Y_i \sim \bernoulli{}(\mu_i)$.
We consider two different distributions for $X$.
First, we generate $X \sim \normal(0, I_p)$, as studied in the theory.
Second, following an experiment in \citet{SurCa2019}, we draw $X$ from a discrete distribution that models feature distributions encountered in Genome-Wide Association Studies (GWAS). 
Here, each feature $X_{ij}$ takes values in $\{0, 1, 2\}$, according to the Hardy-Weinberg equilibrium with parameter $p_j$ varying between covariates in the range $[0.25, 0.75]$.
This setting is not explicitly covered by the theory, but we expect the approximation to work well nonetheless because the features have thin tails.

\begin{figure*}[t]
\begin{center}
\begin{minipage}[t]{0.68\textwidth}
\begin{subfigure}[t]{0.48\textwidth}
\includegraphics[width=\textwidth]{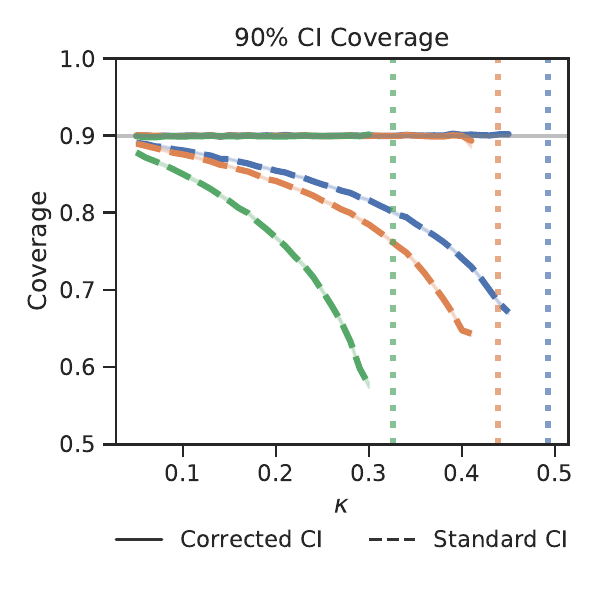}
\caption{Gaussian features}
\label{fig:coverage}
\end{subfigure}
\begin{subfigure}[t]{0.48\textwidth}
\includegraphics[width=\textwidth]{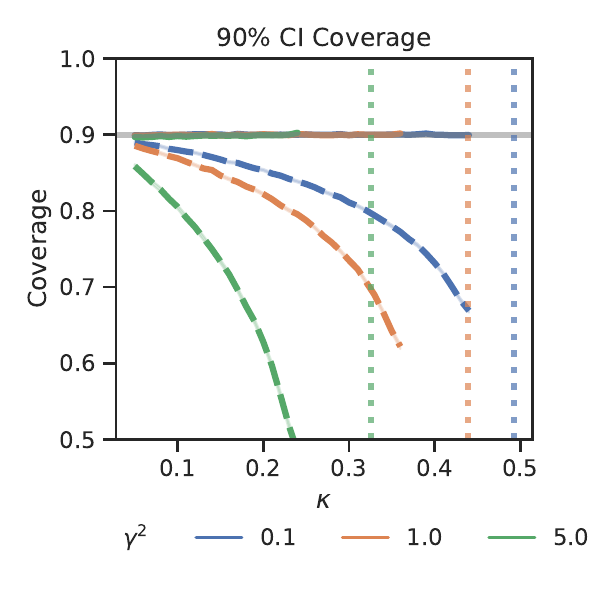}
\caption{Simulated GWAS features}
\label{fig:coverage-gwas}
\end{subfigure}
\caption{Coverage of true probabilities with 90\% standard and corrected CIs over all test examples and over $1000$ simulations, with $n=4000$. Curves are terminated when the observed data are separable in more than 10\% of simulations and simulations with numerical instabilities near the separable frontier (vertical dotted lines) are dropped.}
\end{minipage}%
\begin{minipage}[t]{0.32\textwidth}
\centerline{\includegraphics[width=\textwidth]{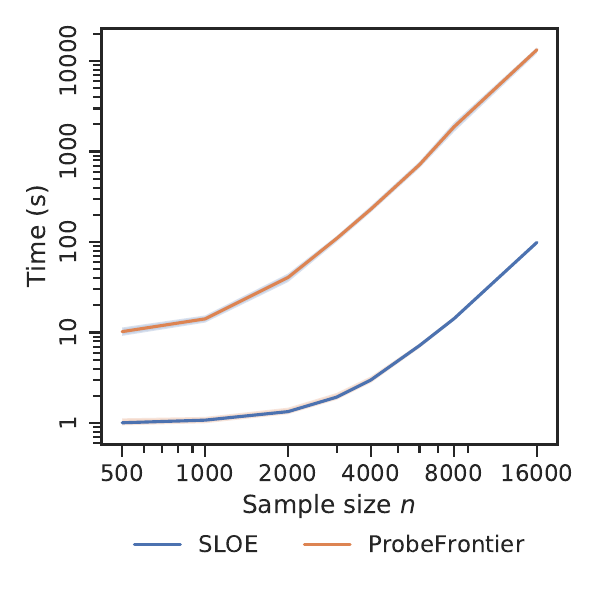}}
\caption{Comparison of wall clock run time for parameter and variance estimation in ProbeFrontier and SLOE over $100$ simulations with $\kappa = 0.2$ and $\gamma^2 = 1$ at various sample sizes.}
\label{fig:runtime}
\end{minipage}
\end{center}
\vspace{-.2in}
\end{figure*}

We show the results of these coverage experiments, with $n=4000$, in Figure~\ref{fig:coverage}.
In both the Gaussian and simulated GWAS settings, across all practically feasible
values of $\kappa$ and $\gamma$, the corrected intervals computed with $\what \eta_{\SLOE}$ reliably provide 90\% coverage.
This is in stark contrast to CIs generated under standard theory whose coverage begins to drop off at $\kappa > 0.05$ (that is, $20$ observations per covariate) for all signal strengths.
Notably, in the simulated GWAS setting that is not covered by theory, the corrected intervals remain reliable while the standard CIs perform even worse than in the Gaussian setting.


SLOE is accurate in finite samples and computationally efficient.
Figure~\ref{fig:runtime} compares the time to perform estimation and inference with both SLOE and ProbeFrontier. For all dataset sizes, SLOE is 10x faster, and for $n \gtrsim 3000$, SLOE is 100x faster, which is when speed matters the most.


\section{Applications}

Here, we present applications of dimensionality corrections with $\what \eta_{\SLOE}$, emphasizing the real-world impact of the differences between standard and corrected statistical inference in logistic regression. 

\subsection{Prediction of Heart Disease}

\begin{figure*}[t]
\begin{center}
	\begin{subfigure}[c]{0.47\columnwidth}
		\centering
		\includegraphics[width=\columnwidth]{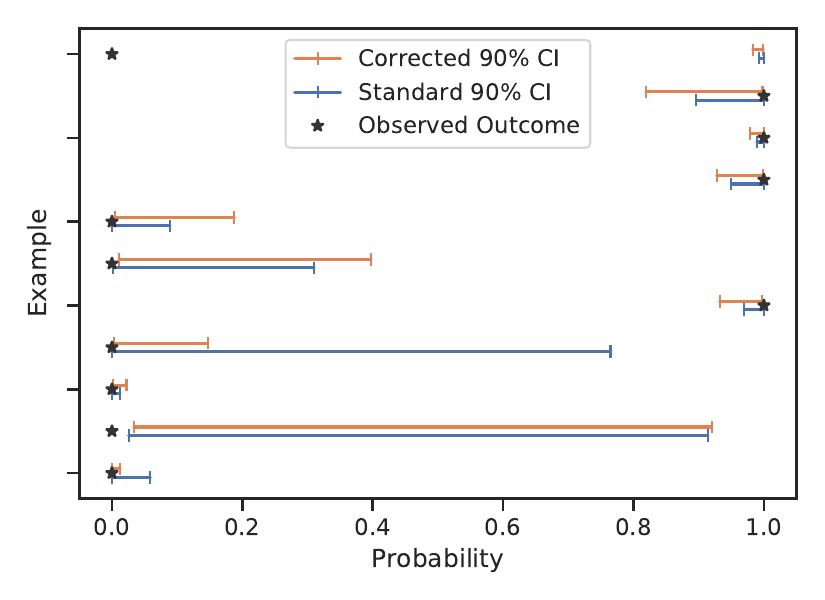}
	\end{subfigure}
	\quad
	\begin{subfigure}[c]{0.47\columnwidth}
		\centering
		\includegraphics[width=\columnwidth]{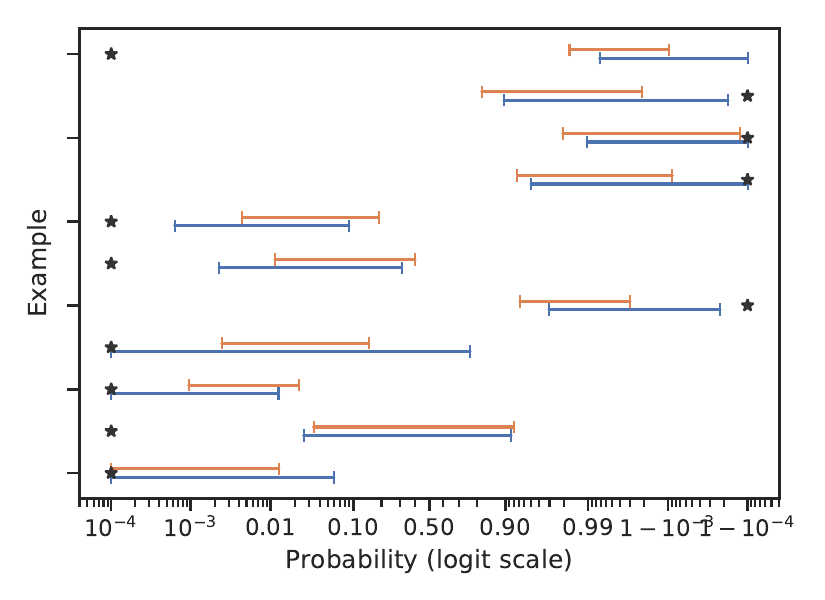}
	\end{subfigure}
\caption{Predictions and 90\% confidence intervals for 8 held-out examples from the Cleveland Clinic Heart Disease dataset, and the observed outcomes, shown with the probabilities on the (a) absolute scale, and (b) logit scale.}
\vspace{-1em}
\label{fig:heart-preds}
\end{center}
\end{figure*}

Reliable uncertainty quantification is essential for supporting effective decision-making in clinical applications \citep{KompaSnBe21}.
Here, we show how the uncertainty communicated by CIs changes substantially based on dimensionality corrections in predicting a heart disease diagnosis from physiological and demographic information from the Cleveland Clinic Heart Disease dataset \citep{DetranoJaStPfScSaGuLeFr89}.
\citet{KompaSnBe21} used this dataset to demonstrate the importance of uncertainty estimation, by showing that logistic regression predictions are sensitive to bootstrapping the training data.
\footnote{The nonparametric bootstrap turns out to produce invalid CIs when $\kappa > 0$; see the simulation in the Supplementary Materials.}

The Heart Disease dataset (downloadable from the UCI Machine Learning Repository) has 14 predictors, and 303 observations. Converting categorical variables to dummy variables, taking interactions between sex and pain categorization given its clinical significance \citep{MehtaBeElVaQuPeBa19}, and balancing the classes, gives 136 training examples and 20 predictors ($\kappa = 0.15$).
SLOE estimates a bias inflation factor $\what{\alpha}$ of $1.40$ for this data, so the logits in the standard MLE will be inflated by a factor of 40\%. Correcting these inflated logits significantly changes the center of the CIs for downstream predictions. Figure~\ref{fig:heart-preds} compares the standard CIs to the corrected CIs for a handful of test examples, along with the true labels. Notice that the standard CIs are overconfident, with the top CI being entirely above $99.8\%$ probability of heart disease, when in fact, there wasn't. The corrected CIs show that in reality, much higher probabilities of misdiagnosis (i.e., 1 in 58 chance of being wrong, rather than 1 in 140) are consistent with the training data. In clinical settings, the uncorrected CIs could communicate an unwarranted sense of confidence in a diagnosis.

\subsection{Genomics}

\begin{figure*}[!tpb]
\centering

	\begin{subfigure}[t]{0.6\linewidth}
		\centering
		\includegraphics[width=\columnwidth]{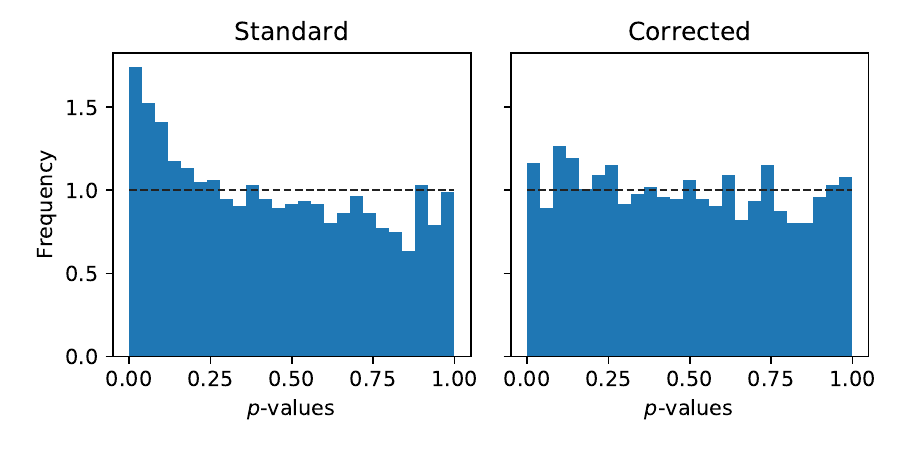}
	\end{subfigure}
	\quad
	\begin{subfigure}[t]{0.3\linewidth}
		\centering
		\includegraphics[width=\columnwidth]{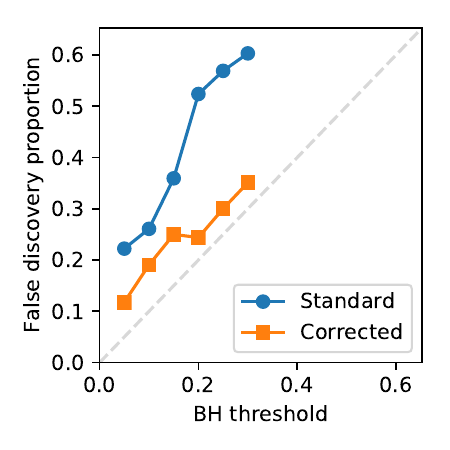}
	\end{subfigure}

\caption{(a) Histogram of $p$-values for null coefficients (randomly selected SNPs not known to be associated with glaucoma), from the standard and corrected logistic regression models. (b) Comparison of the Benjamini–Hochberg Q-value threshold and the empirical proportion of false discoveries, using the $p$-values from the standard and corrected logistic regression models.}
\vspace{-1em}
\label{fig:gwas-p-values}

\end{figure*}

In medical genomics, it is common to build predictors for clinically relevant binary phenotypes with logistic regression models, using genetic variants as features. Practitioners often use $p$-values associated with coefficients to interpret these models (e.g. \citealp{Ding2020.11.30.403188}). Here, we show in a real example that the corrected $p$-values computed with $\what \eta_{\SLOE}$ maintain their intended interpretation, unlike the standard $p$-values.

Specifically, we build a model that predicts glaucoma, an eye condition that damages the optic nerve, on the basis of genetic predictors. We build a dataset from UK Biobank \citep{Sudlow2015-mr} with 111 variants directly associated with glaucoma according to a recent meta-analysis of GWAS~\citep{Gharahkhani2020.01.30.927822} and randomly sample 1744 ``null'' single nucleotide polymorphisms (SNPs) also present in HapMap 3~\citep{International_HapMap_3_Consortium2010-yv}, such that each sampled SNP is at least 1M base pairs away from all associated variants and from each other to avoid correlated features due to linkage disequilibrium. This generates 1855 genetic predictors in total. 
Glaucoma status was determined based on touchscreen questionnaire responses and hospital episode statistics as performed previously~\citep{Khawaja2018-fq,alipanahi2020largescale}.

From this, we construct a training set of 5067 cases with glaucoma and 5067 age- and sex-matched controls ($\kappa = 1855 / 10134 = 0.183$), \footnote{We reserve 1126 individuals (563 cases, 563 controls) sampled in the same way for a test set.} train a logistic regression model, and compute standard and corrected $p$-values with it.

Figure~\ref{fig:gwas-p-values}a shows $p$-values of the randomly selected variants whose effects we expect to be null. The standard $p$-values are skewed while the corrected $p$-values follow the expected uniform distribution closely, as we observed in simulated data (Figure~\ref{fig:pvalues}).
Uniform null $p$-values are important for controlling the error rate in procedures applied to select important variants, e.g., for resource-intensive follow-up. 
In Figure~\ref{fig:gwas-p-values}b, we investigate how these $p$-values interact with the  Benjamini-Hochberg (BH) procedure~\citep{BenjaminiHo95} for ``discovering'' important variants while controlling false discovery rates.
To compute an approximate empirical false discovery proportion, we take all variants identified by the BH procedure, and compute the proportion of the null variants (as opposed to known glaucoma-associated variants) in this set.
The corrected $p$-values result in better false discovery proportion calibration across a range of false discovery rate targets (Figure~\ref{fig:gwas-p-values}b).

\section{Discussion}
The groundbreaking work by \citet{SurCa2019} has significant implications for theoretical understanding, uncertainty quantification, and generally, practical statistical inference in large scale data.
Connecting this theory to practice highlights some interesting insights and questions.

\paragraph{Bias and Variance}
In the regime that we analyze, these dimensionality corrections constitute a something of a statistical free lunch: the corrections reduce both bias and variance, or, in the case of confidence intervals, they both improve coverage to their nominal level (Figure~\ref{fig:coverage}) while shortening their length on the logit scale (Figure~\ref{fig:heart-preds}b).
This occurs because bias is created by overfitting\footnote{To overfit in this logistic regression setting, the model reduces training residuals by amplifying coefficient to produce extreme predictions.}, as opposed to underfitting, which we can correct. This systematic bias in the parameters appears in downstream predictions, on the logit and probability scale, and unless corrected, does not vanish asymptotically.
Because the bias correction shrinks the coefficient estimates by a factor of $1/\alpha$, it also produces lower-variance predictions, even after correcting for the too-small variance predicted by standard theory.

\paragraph{CIs and Regularized Predictions}
CIs with valid coverage are an attractive tool for quantifying uncertainty about predictions, but in practice, point predictions are usually made with models that are regularized to reduce the variance of the predictions, at the cost of some amount of bias.
Yet, the CIs around the predictions that we produce are centered around unbiased estimates of the logits; thus, one might ask whether these can produce coherent summaries together.
In simulations, we find that the $1-\delta$ CIs contain the optimal regularized predictor (selected via leave-one-out cross validation) more than a $1-\delta$ fraction of the time. This makes sense, as the LOOCV regularized predictor should be closer to the true probability, which we know is contained within the CIs $1-\delta$ of the time. If one preferred the CIs to always include the regularized predictions, taking the convex hull of the CIs and regularized predictions would give slightly larger, but more interpretable, CIs with higher-than-nominal coverage.

\paragraph{Broader Implications for ML}
These observations potentially have large implications in machine learning. The logistic regression MLE is closely related to the softmax loss used frequently in deep learning. However, the paradigm studied here focuses on the underparameterized regime, whereas many deep learning models are overparameterized. Even in large datasets where this may not be true, the signal is often strong enough that that the data are separable, and so the MLE does not exist. \citet{SalehiAbHa19} characterizes generalization into regimes where the data are separable by adding regularization, but their results rely more on assuming isotropic Gaussian features, whereas this work and \citet{ZhaoSuCa2020} generalize well theoretically to elliptical Gaussian features and experimentally to sub-Gaussian features.

\section*{Acknowledgements}
This research has been conducted using the UK Biobank Resource under Application Number 17643. The authors thank Babak Alipanahi and Farhad Hormozdiari for their help and feedback on the genomics experiments, Arjun Seshadri for his feedback on the theoretical results and advice on linear programming to implement ProbeFrontier efficiently, and D. Sculley for his support and feedback on the research.

\bibliographystyle{abbrvnat}
\bibliography{bib}

\appendix

\section{Bootstrap Confidence Intervals}
Here, we provide additional information about the confidence intervals presented in the main text in Figure~\ref{fig:coverage}, and additionally include bootstrap CIs. We used the nonparametric multiplier bootstrap \citep{Praestgaard1990}, where in each bootstrap sample, the MLE is refit with each example weighted by an iid Poisson distribution with rate parameter $\lambda = 1$. This very closely approximates the nonparametric bootstrap with sampling with replacement. For each test prediction, the $1-\delta$ CIs are calculated using the $\delta / 2$ and $1-\delta/2$ quantiles of the bootstrap estimates, known as the percentile bootstrap \citep{EfronTi93}.
\begin{center}
    \centering
    \includegraphics[width=0.75\textwidth]{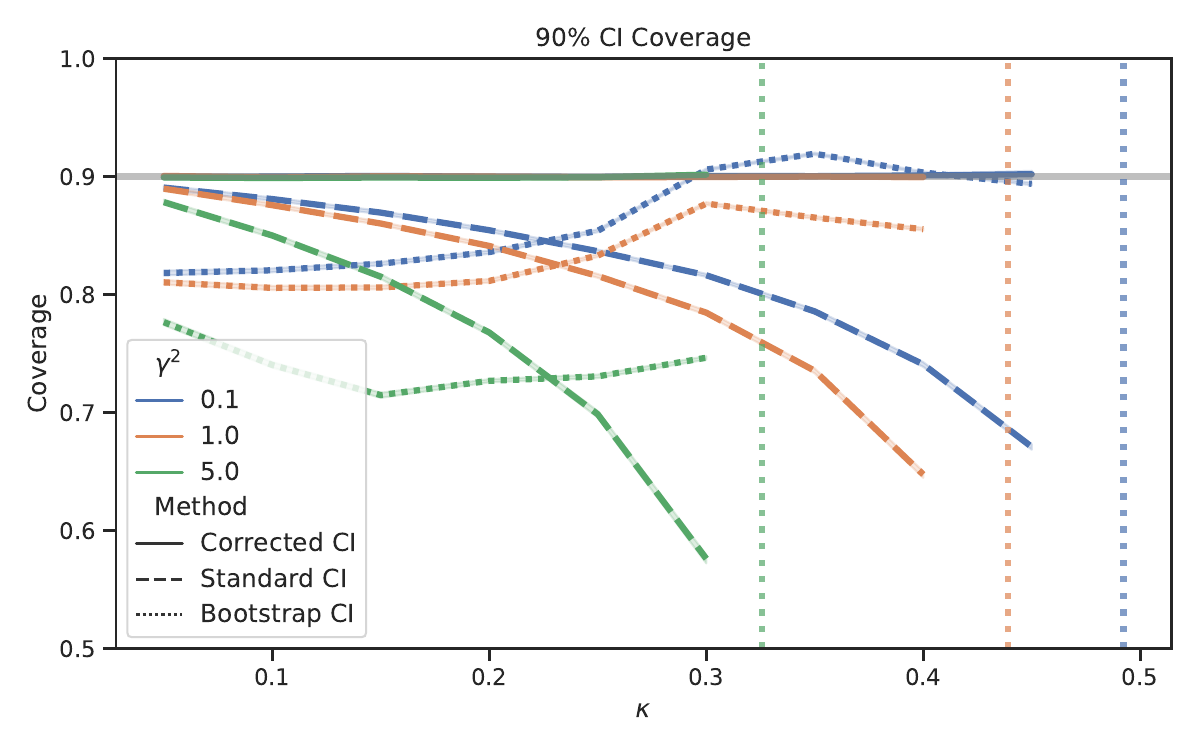}
    \captionof{figure}{Same setting as main Figure~\ref{fig:coverage}, including bootstrapped CIs. Notice that the bootstrap CIs do not have proper coverage.}
    \vspace{2em}
    \label{fig:bootstrap-coverage}
    \includegraphics[width=0.75\textwidth]{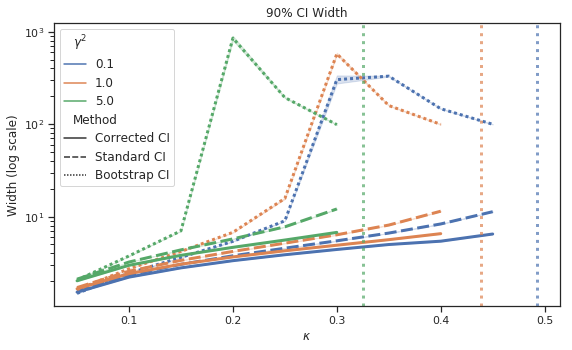}
    \captionof{figure}{Same setting as main Figure~\ref{fig:coverage}, but compares the width of the confidence intervals. Notice the large bootstrap CIs, especially when they approach or exceed nominal coverage.}
    \label{fig:bootstrap-width}
    \vspace{2em}
    \centering
    \includegraphics[width=0.75\textwidth]{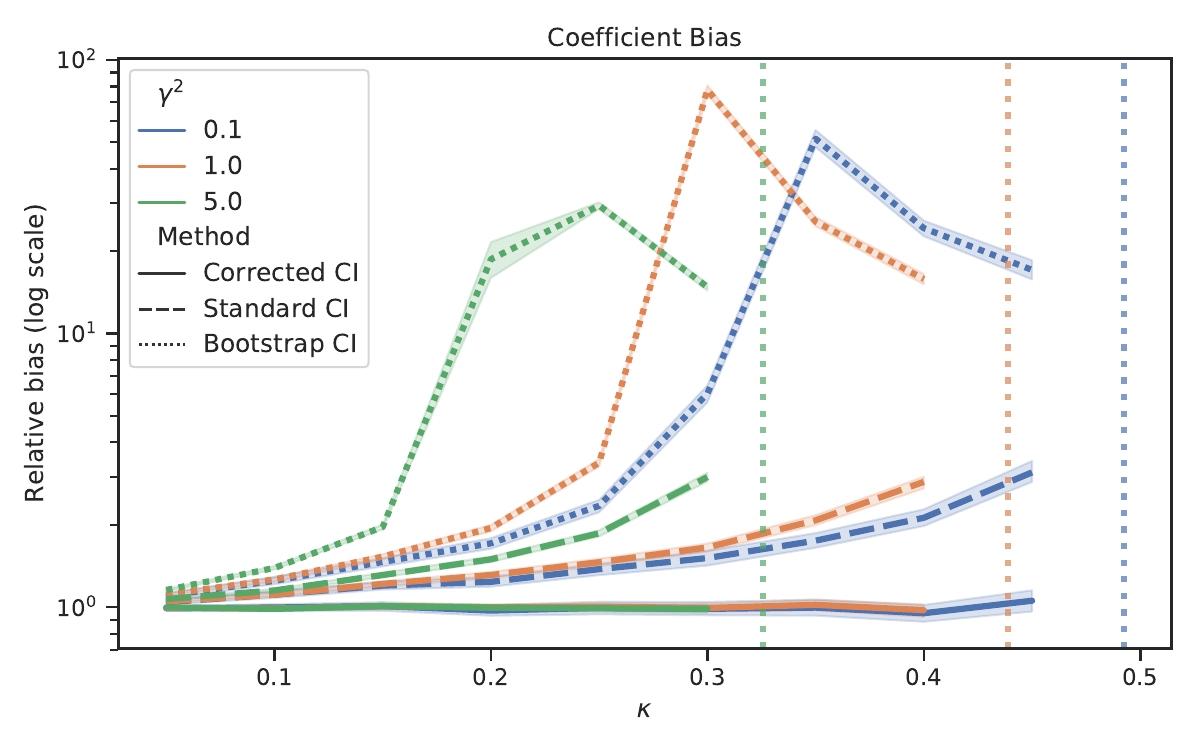}
    \captionof{figure}{Same setting as main Figure~\ref{fig:coverage}, but compares the magnitude of the logits of the predictions relative to the magnitude of the true logit, for any test example with $|x^\top \beta| > 0.01$. For the bootstrap, this is calculated from the median estimate over bootstrap samples, throwing out any where the data were separable and so the estimate is at $\pm \infty$.}
    \label{fig:bootstrap-bias}
\end{center}

\section{Proofs}
\label{sec:proofs}
Throughout the proofs, we will use $C > 0$ as a generic constant that could always be made larger without invalidating a statement, and $c > 0$ a generic constant that could always be made smaller without invalidating a statement. This way, we can avoid reducing the readability due to onerous constant accounting.


\begin{proof-of-theorem}[\ref{thm:consistency}]
First, we show that without loss of generality, we can consider the case where $\Sigma = I_p$, the $p \times p$-dimensional identity matrix. To see that this is sufficient, we apply Proposition 2.1 of \citet{ZhaoSuCa2020}, which states the following.

\begin{proposition}[Proposition 2.1, \citet{ZhaoSuCa2020}]
Fix any matrix $L$ obeying $\Sigma = LL^\top$, and consider the vectors
\begin{equation*}
    \what{\theta} = L^\top \what{\beta},~\mbox{and}~\theta = L^\top \beta.
\end{equation*}
Then, $\what{\theta}$ is the MLE in a logistic model with regression coefficient $\theta$ and covariates drawn i.i.d. from $\normal{}(0, I_p).$
\end{proposition}
With this in mind, we can prove everything in this rotated setting, and as long as $L^\top$ is full rank, the results (up to appropriate scaling) will hold for $\what{\beta}$, as well. Choosing $L^\top$ to be a Cholesky decomposition, it will satisfy $LL^\top = \Sigma$, and will be full rank with a bounded operator norm for $L$ and its inverse $L^{-1}$, because of the assumption that the condition number of $\Sigma$ is bounded.

Now, we prove the result under the assumption that $\Sigma = I_p$ in three steps. The first is to show that $\|\what{\beta}\|_2^2 \cp \eta$. The second is to show that the leave-one-out estimators are close enough in norm to $\what\beta$ such that
\begin{equation*}
   \frac{1}{n}\sum_{i=1}^n \|\what{\beta}_{-i}\|_2^2 - \|\what{\beta}\|_2^2 \cp 0.
\end{equation*}
The third is to show that the estimator $\what{\eta}_{\text{LOO}}^2$, which incorporates an empirical estimate of $\Sigma$, concentrates around $\bar{\eta}^2 = \frac{1}{n}\sum_{i=1}^n \|\what{\beta}_{-i}\|_2^2$.

The first step is a direct application of Theorem 2 of \citet{SurCa2019}, which we restate here using our problem scaling.
\begin{theorem}[Theorem 2, \citet{SurCa2019}]
Assume the dimensionality and signal strength parameters $\kappa$ and $\gamma$ are such that $\gamma < g_{\text{MLE}}(\kappa)$ (the region where the MLE exists asymptotically and is shown in \citet[Fig. 6]{SurCa2019}). Assume the logistic model described where the empirical distribution of $\{\sqrt{n} \beta_j\}$ converges weakly to a distribution $\Pi$ with finite second moment. Suppose further that the second moment converges in the sense that as $n \to \infty$, $\text{Ave}_j(n \beta_j^2) \to E[\beta^2],~\beta\sim \Pi$. Then for any pseudo-Lipschitz function $\psi$ of order 2, the marginal distribution of the MLE coordinates obeys
\begin{equation*}
    \frac{1}{p} \sum_{j=1}^p \psi(\sqrt{n}(\what{\beta}_j - \alpha_\star \beta_j, \beta_j) \cas \E[\psi(\sigma_{\star} Z, \beta)],~~Z \sim \normal{}(0, 1),
\end{equation*}
where $\beta \sim \Pi$, independent of $Z$.
\end{theorem}

The first step follows from this theorem using $\psi(t, u) = (t + \alpha_\star u)^2$, because $\gamma^2 = \var(\beta^\top X) = \|\beta\|_2^2 = \E_\Pi[\beta^2]$.

The second step involves showing that the difference in norms between $\what{\beta}$ and $\what{\beta}_{-i}$ is small with high probability. To do so, we use the following lemma. In this lemma, and throughout the proofs, we will use sequences $K_n$ and $H_n$, satisfying the following conditions:
for any $c_H, c_K, \epsilon > 0$,
\begin{equation}
\label{eq:sequence-cond}
    K_n = o(n^{\epsilon}),~H_n = o(n^{\epsilon}),~n^8 \exp(-c_H H_n^2) = o(1),~\mbox{and}~n^7 \exp(-c_K K_n^2) = o(1).
\end{equation}
Taking $H_n = K_n = \log n$, for example, would satisfy these conditions.

\begin{lemma}
\label{lem:loo-norm-diff}
Let $\what{\beta}$ be the MLE and $\what{\beta}_{-i}$ be the MLE excluding the $i$-th example. Let $K_n$ and $H_n$ be sequences satisfying \eqref{eq:sequence-cond}. Then, there exists universal constants $C, c > 0$ such that
\begin{align*}
    P\left( \left\| \what{\beta}_{-i} - \what\beta \right\| > C\left(C\frac{K_n}{\sqrt{n}} + \frac{K_n^2 H_n}{n}\right) \right) &\le \exp(-\Omega(n)) + C \exp(-c K_n^2)\\
    &~~~~~~+ C n \exp( -c H_n^2) + C \exp(-c n(1 + o(1))).
\end{align*}
A similar bound holds for $\|\what{\beta}_{-i} - \what{\beta}_{-ik}\|$, where $\what{\beta}_{-ik}$ is the MLE with the $i$-th and $k$-th example excluded.
\end{lemma}
\noindent See Section~\ref{sec:lemma-proofs} for proof.
Taking a union bound over the probabilities bounded in Lemma~\ref{lem:loo-norm-diff} for each $i$ gives that the event $G = \left\{ \sup_{i} \left\| \what{\beta}_{-i} - \what\beta \right\| \le C_1\left(\frac{K_n}{\sqrt{n}} + \frac{K_n^2 H_n}{\sqrt{n}}\right) \right\}$ is bounded with probability $1-o(1)$ using the conditions \eqref{eq:sequence-cond}. Therefore, conditional on $G$, we can write
\begin{align*}
   \left| \frac{1}{n}\sum_{i=1}^n \|\what{\beta}_{-i}\|_2^2 - \|\what{\beta}\|_2^2 \right| &\le \frac{1}{n}\sum_{i=1}^n \frac{C_1^2 K_n^4 H_n^2}{n} = \frac{C_1^2 K_n^4 H_n^2}{n} \to 0,
\end{align*}
with probability converging to $1$.

Our strategy for the third step is to show that $\E[ \what{\eta}^2 - \bar{\eta}^2] = 0$, and then apply Chebyshev's inequality and bound the variance. The challenge is that the terms in $\what{\eta}_{\text{LOO}}^2$ are not independent, and therefore, we will need to show that the covariances are asymptotically negligible. To do so, we will employ a leave-\emph{two}-out argument, inspired by the proof techniques of \citet{ElKaroui18} and \citet{SurCa2019}.

First, write $\what{\eta}^2_{\text{LOO}} - \bar{\eta}^2$ as
\begin{equation*}
    \frac{1}{n}\sum_{i=1}^n \what{\beta}_{-i}^\top \left(X_i X_i^\top - I\right) \what{\beta}_{-i}.
\end{equation*}
Using that $\what{\beta}_{-i}$ is independent of $X_i$, we immediately can conclude that $\E[ \what{\eta}^2 - \bar{\eta}^2] = 0.$

Next, we will use Chebyshev's inequality to bound the probability using the variance. To show that the variance goes to zero, we will need to show that, for $i \neq k$, the covariance between terms
\begin{equation*}
    \what{\beta}_{-i}^\top (X_i X_i^\top - I) \what{\beta}_{-i} ~~\mbox{and}~~\what{\beta}_{-k}^\top (X_k X_k^\top - I) \what{\beta}_{-k}
\end{equation*}
converges to zero. The challenge in doing so is that the MLE in one term is dependent on the covariate $X$ in the other term. We solve this challenge by showing the estimated coefficients $\what{\beta}_{-i}$ and $\what{\beta}_{-k}$ are close enough (for our purposes) to $\what{\beta}_{-ik},$ the MLE with both the $i$-th and $k$-th predictor excluded. Consider the following sequence of events,
\begin{equation}
    \label{eq:lto-prods}
    E_n = \left\{ \sup_{i} \sup_{k \not= i} |X_i^\top (\what{\beta}_{-ik} - \what{\beta}_{-i})| \le \frac{C K_n^2 H_n}{\sqrt{n}}, \sup_{i} |X_i^\top \what{\beta}_{-i}| \le C \right\}.
\end{equation}
The following lemma show that this sequence of events has probability approaching $1$.

\begin{lemma}
\label{lem:sup-loo}
Let $\what{\beta}_{-i}$ be the MLE with the $i$-th example held out, and $\what{\beta}_{-ij}$ be the MLE with the $i$-th and $j$-th examples held out. Then, there exists universal constants $C, c > 0$ such that
\begin{gather*}
    P\left( \sup_{i} \sup_{k \not= i} |X_i^\top (\what{\beta}_{-ik} - \what{\beta}_{-i})| \le \frac{C K_n^2 H_n}{\sqrt{n}} \right) \ge 1 - C n^2 \exp(-c H_n^2)\\
    \quad\,\quad\,\quad - C n \exp(-c K_n^2) - C n \exp(-c n(1 + o(1))),
\end{gather*}
and
\begin{gather*}
    P\left( \sup_{i} |X_i^\top \what{\beta}_{-i}| \le C \right) \ge 1 - C n^2 \exp(-c H_n^2)\\
    \quad\,\quad\,\quad - C n \exp(-c K_n^2) - C n \exp(-c n(1 + o(1))) - C n \exp(-\Omega(n)),
\end{gather*}
\end{lemma}
\noindent As before, the proof of this lemma is in Section~\ref{sec:lemma-proofs}.

Additionally, we will need to control the norm of the difference between the leave-one-out and leave-two-out estimators. Let
\begin{equation*}
    \label{eq:loo-norms}
    B_n = \left\{ \sup_{i} \|\what{\beta}_{-i}\|_2 \le C,~\sup_{i} \sup_{k \not= i} \|\what{\beta}_{-ik} - \what{\beta}_{-i}\|_2 \le \frac{C (K_n + K_n^2 H_n)}{\sqrt{n}} \right\}.
\end{equation*}
Writing this as the intersection of the events $B_{ik} =  \{ \|\what{\beta}_{-i}\|_2 \le C,~\|\what{\beta}_{-ik} - \what{\beta}_{-i}\|_2 \le \frac{C (K_n + K_n^2 H_n)}{\sqrt{n}} \}$, a union bound over the complements $B_n^C = \bigcup_{i \not= j} B_{ij}^C$, along with the control on the probability $P(B_{ij}^C)$ implied by \citet[Theorem 4]{SurChCa19} and Lemma~\ref{lem:loo-norm-diff}, respectively, shows that this sequence of event $(B_n)_{n=1}^\infty$ has probability approaching $1$.

Now, we proceed with bounding the probability that $\what{\eta}$ is far from $\bar{\eta}$. Let $\epsilon > 0$.
\begin{align*}
    P\left( |\what{\eta}^2_{\text{LOO}} - \bar{\eta}^2| > \epsilon \right) \le P\left( |\what{\eta}_{\text{LOO}}^2 - \bar{\eta}^2| > \epsilon \mid B_n \cap E_n \right) + P(E_n^{C} \cup B_n^{C}).
\end{align*}
Lemma~\ref{lem:sup-loo} shows that $\lim_{n\to\infty} P(E_n^{C}) = 0$. Above, we showed that $\lim_{n\to\infty} P(B_n^{C}) = 0$, and so $\lim_{n\to\infty} P(E_n^{C} \cup B_n^{C}) = 0$. Therefore, what remains is to control $P\left( |\what{\eta}^2_{\text{LOO}} - \bar{\eta}^2| > \epsilon \mid B_n \cap E_n \right)$.

For notational convenience, denote
\begin{align*}
    \goodP(\cdot) &\defeq P(\cdot \mid B_n \cap E_n),\\
    \goodE[\cdot] &\defeq E[\cdot \mid B_n \cap E_n],~\text{and}\\
    \goodvar(\cdot) &\defeq \var(\cdot \mid B_n \cap E_n).\\
\end{align*}

Applying Chebyshev's inequality,
\begin{equation*}
    \goodP\left( |\what{\eta}_{\text{LOO}}^2 - \bar{\eta}^2| > \epsilon \right) \le \frac{\goodvar\left( \frac{1}{n}\sum_{i=1}^n \what{\beta}_{-i}^\top \left(X_i X_i^\top - I\right) \what{\beta}_{-i} \right)}{\epsilon^2}.
\end{equation*}
Showing that $\goodvar\left( \frac{1}{n}\sum_{i=1}^n \what{\beta}_{-i}^\top \left(X_i X_i^\top - I\right) \what{\beta}_{-i} \right) \to 0$ completes the proof. To do so, expand the sum as
\begin{align*}
    & \goodvar\left( \frac{1}{n}\sum_{i=1}^n \what{\beta}_{-i}^\top \left(X_i X_i^\top - I\right) \what{\beta}_{-i} \right)
    \\
    &~~~~= \frac{1}{n^2}\left( \sum_{i=1}^n \goodvar(\what{\beta}_{-i}^\top \left(X_i X_i^\top -  I\right) \what{\beta}_{-i}) \right. \\
    &~~~~~~~~~~~~~\left. + \sum_{i \not= k} \goodE \left[     \what{\beta}_{-i}^\top (X_i X_i^\top -  I) \what{\beta}_{-i} \what{\beta}_{-k}^\top (X_k X_k^\top - I) \what{\beta}_{-k} \right] \right)
\end{align*}
On $B_n$, $\what{\beta}_{-i}$ all have bounded norm. The known normal distribution of the $X_i$ allows us to conclude that the $n$ variance terms $\goodvar(\what{\beta}_{-i}^\top \left(X_i X_i^\top - I\right) \what{\beta}_{-i})$ will be bounded by some fixed constant. What remains is to control the $n(n-1)$ covariance terms.

Consider the $i$-th and $k$-th covariance term,
\begin{align*}
    \goodE \left[     \what{\beta}_{-i}^\top (X_i X_i^\top - I) \what{\beta}_{-i} \cdot \what{\beta}_{-k}^\top (X_k X_k^\top - I) \what{\beta}_{-k} \right].
\end{align*}
\noindent The challenge is that $\what{\beta}_{-i}$ is not independent of $X_k$, which prevents us from splitting these terms into the product of their expectations. With this in mind, we imagine instead that the first quadratic form was $ \what{\beta}_{-ik}^\top (X_i X_i^\top - I) \what{\beta}_{-ik} $, which would be independent of $X_k$, and then study the remainder terms. For notational convenience, denote 
\begin{equation*}
    Z_i = X_iX_i^\top - I.
\end{equation*}
Writing $\what{\beta}_{-i} = \what{\beta}_{-ik} + \what{\beta}_{-i} - \what{\beta}_{-ik}$, the above covariance expands as
\begin{align}
&\goodE \left[     (\what{\beta}_{-ik} + \what{\beta}_{-i} - \what{\beta}_{-ik})^\top Z_i (\what{\beta}_{-ik} + \what{\beta}_{-i} - \what{\beta}_{-ik}) \what{\beta}_{-k}^\top Z_k \what{\beta}_{-k} \right] \nonumber
\\
&~~~~~~= 
\goodE \left[   \left(  \what{\beta}_{-ik}^\top Z_i \what{\beta}_{-ik} + 2(\what{\beta}_{-i} - \what{\beta}_{-ik}) Z_i \what{\beta}_{-ik} + (\what{\beta}_{-i} - \what{\beta}_{-ik})^\top Z_i (\what{\beta}_{-i} - \what{\beta}_{-ik})\right) \what{\beta}_{-k}^\top Z_k \what{\beta}_{-k} \right] \nonumber
\\
&~~~~~~= 
\goodE \left[ \what{\beta}_{-ik}^\top Z_i \what{\beta}_{-ik} \what{\beta}_{-k}^\top Z_k \what{\beta}_{-k} \right] \label{eq:cov-expansion}
\\
&~~~~~~~~~~~~+
\goodE \left[   \left(2(\what{\beta}_{-i} - \what{\beta}_{-ik})^\top Z_i \what{\beta}_{-ik} + (\what{\beta}_{-i} - \what{\beta}_{-ik})^\top Z_i (\what{\beta}_{-i} - \what{\beta}_{-ik})\right) \what{\beta}_{-k}^\top Z_k \what{\beta}_{-k} \right] \nonumber
\end{align}
Because $\E[Z_k \mid \{(X_i, Y_i)\}_{i\not=k}] = 0$, we expect that the first term of \eqref{eq:cov-expansion} should be nearly $0$, as well, except that we have conditioned on $B_n \cap E_n$, which might change the distribution of $Z_k$. The following lemma controls the difference between $\E[ Z_k \mid \{X_i, Y_i)\}_{i\not=k}]$ and $\goodE[ Z_k \mid \{X_i, Y_i)\}_{i\not=k}]$, showing that it vanishes asymptotically.

\begin{lemma}
    Let $E_n$ be defined as in \eqref{eq:lto-prods} and $B_n$ as in \eqref{eq:loo-norms}. Under the conditions of Theorem~\ref{thm:consistency},
    \begin{equation}
      \left| \goodE \left[ \what{\beta}_{-ik}^\top Z_i \what{\beta}_{-ik} \what{\beta}_{-k}^\top Z_k \what{\beta}_{-k} \right]\right| = o\left( \frac{1}{n} \right).
      \label{eq:expectation-nearly-zero}
    \end{equation}
    \label{lem:expectation-nearly-zero}
\end{lemma}

We bound the remaining terms by using the properties of the events $B_n$ and $E_n$, on which we've conditioned.
\begin{align*}
    &\left|\left(2(\what{\beta}_{-i} - \what{\beta}_{-ik}) Z_i \what{\beta}_{-ik} + (\what{\beta}_{-i} - \what{\beta}_{-ik})^\top Z_i (\what{\beta}_{-i} - \what{\beta}_{-ik})\right) \what{\beta}_{-k}^\top Z_k \what{\beta}_{-k}\right|\\
    &~~~~~~= \left|2(\what{\beta}_{-i} - \what{\beta}_{-ik})^{\top} Z_i \what{\beta}_{-ik} + (\what{\beta}_{-i} - \what{\beta}_{-ik})^\top Z_i (\what{\beta}_{-i} - \what{\beta}_{-ik})\right| \cdot |\what{\beta}_{-k}^\top Z_k \what{\beta}_{-k}|
\end{align*}
The second term is bounded on the event $B_n$, using the bounded norm of $\what{\beta}_{-k}$ and the fact that $X_i$ is normally distributed with variance $I_p$. Conditional on the set $E_n \cap B_n$, the first term is bounded by $\sqrt{\frac{C^3 K_n^4 H_n^2}{n}} + \frac{CK_n^4H_n^2}{n}$, for some universal constant $C > 0$.

Plugging all of these into the expression for the variance, we see that
\begin{equation*}
    \goodvar\left( \frac{1}{n}\sum_{i=1}^n \what{\beta}_{-i}^\top \left(X_i X_i^\top - I\right) \what{\beta}_{-i} \right) \lesssim \frac{K_n^2 H_n}{\sqrt{n}} + \frac{1}{n},
\end{equation*}
which shows that
\begin{equation*}
    \lim_{n\to \infty} P(|\what{\eta}^2 - \bar{\eta}^2| > \epsilon) = 0,
\end{equation*}
concluding the proof of Theorem~\ref{thm:consistency}.
\end{proof-of-theorem}

\subsection{Approximation Error of Taylor Expansion}
\label{sec:taylor-precise}

\begin{proof-of-proposition}[\ref{prop:SLOE-loo}]
First, we derive the remainder between $\what{\beta}_{-i}$ and $\what{\beta} + H_{-i}^{-1}X_i(Y_i - g(\what{\beta}^\top X_i))$. Then, we show that these remainder terms in the difference between $\what{\eta}_{\SLOE}$ and $\what{\eta}_{\text{LOO}}$ vanish.

Starting from Eq.~\eqref{eq:diff-of-est-eqs}, we apply a Taylor expansion with the remainder given by the Mean Value Theorem,
\begin{align*}
    X_i(Y_i - g(\what{\beta}^\top X_i)) + \sum_{j \in \mathcal{I}_{-i}} X_j g'(\what{\beta}^\top X_j) X_j^\top (\what{\beta}_{-i}-\what{\beta}) + \sum_{j \in \mathcal{I}_{-i}} X_j \frac{1}{2} g''({\beta^{\circ}_{-i}}^\top X_j)(X_j^\top(\what{\beta}_{-i}-\what{\beta}))^2 = 0,
\end{align*}
for $\beta^{\circ}_{-i} = t \what{\beta} + (1-t) \what{\beta}_{-i}$ for some $t \in [0, 1]$.
Let $R_n = \sum_{j \in \mathcal{I}_{-i}} X_j \frac{1}{2} g''({\beta^{\circ}_{-i}}^\top X_j)(X_j^\top(\what{\beta}_{-i}-\what{\beta}))^2$ be the remainder term that leaves only linear terms. By showing that its norm is growing much more slowly than the other terms in the above equality, we show that it is asymptotically negligible. To do so, for any $0 < \varepsilon < 1/2$, let $V_n = n^{-\varepsilon} R_n$. Then, we have
\begin{align*}
    X_i(Y_i - g(\what{\beta}^\top X_i)) + n^{\varepsilon} V_n + \sum_{j \in \mathcal{I}_{-i}} X_j g'(\what{\beta}^\top X_j) X_j^\top (\what{\beta}_{(-i)}-\what{\beta}) = 0,
\end{align*}
Lemma 17 from \cite{SurChCa19} shows that for $K_n$ and $H_n$ satisfying \eqref{eq:sequence-cond}, $\sup_{i \not= j} |X_j^\top (\what{\beta}_{(-i)} - \what{\beta})| \le C K_n^2H_n / \sqrt{n}$ with probability $1 - \delta_n$ for $\delta_n = Cn \exp(-c H_n^2) - C\exp(-c K_n^2)-\exp(-c n(1 + o(1)))$.
Using the condition~\eqref{eq:sequence-cond} with $\epsilon = \varepsilon / 3$, we know that $n^{-\varepsilon} K_n^2 H_n = o(1)$. Therefore, the above observation along with the fact that $g''(s) \le 1$ for all $s$, implies that $V_n \in \R^d$ satisfies
\begin{equation*}
    \|L^{-1} V_n\|_2^2 \le \frac{C^2}{n} \left\| \frac{1}{\sqrt{n}} \sum_{j \in \mathcal{I}_{-i}} L^{-1} X_j \right\|_2^2
\end{equation*}
with probability at least $1 - \delta_n$, for $L$ a full rank triangular matrix satisfying $LL^\top = \Sigma$. Using that $L^{-1}X_i$ is an isotropic Gaussian, and applying standard concentration bounds for multivariate Gaussians, we get that with probability at least $1 - 2n\exp(-(\sqrt{p} - 1)^2 / 2)$,
\begin{equation*}
    \frac{C^2}{n} \left\| \frac{1}{\sqrt{n}} \sum_{j \in \mathcal{I}_{-i}} L^{-1} X_j \right\|_2^2 \le 2C^2 \kappa.
\end{equation*}
Altogether, $\|L^{-1} V_n\|_2^2 \le 2C^2 \kappa$ with probability at least $1 - \delta_n - 2n\exp(-(\sqrt{p} - 1)^2 / 2)$.

Using this fact about the remainder, we proceed with bounding the difference between $\what{\eta}^2_{\text{LOO}}$ and $\what{\eta}^2_{\SLOE{}}$. For notational convenience, define $\wt{\beta}_{-i} = \what{\beta} + H_{-i}^{-1}X_i(Y_i - g(\what{\beta}^\top X_i))$.
Then,
\begin{align}
    \left| \frac{1}{n}\sum_{i=1}^n (X_i^\top \what{\beta}_{-i})^2 - (X_i^\top \wt{\beta}_{-i})^2 \right| &\le \frac{1}{n} \sum_{i=1}^n (X_i^\top n^\varepsilon H_{-i}^{-1} V_n)^2 \le \frac{1}{n} \sum_{i=1}^n \|X_i\|_2^2 n^{2\varepsilon}\|H_{-i}^{-1} V_n\|_2^2 
    \label{eq:cs-bound}
\end{align}
Standard results for multivariate Gaussians show that $\sup_{i=1,\dots,n} \|X_i\|_2^2 < 4p$ with probability $1 - o(1)$. Therefore, what remains is to bound $\|H^{-1}_{-i} V_n\|_2^2$.

To do so, we take advantage of Lemma 7 from \cite{SurChCa19}, proved in the setting where $X_i \sim \normal{}(0, I_d)$. Therefore, we start by showing that we can convert our problem into one in this setting. Specifically, let $Z_i = L^{-1}X_i$, so that $Z_i \sim \normal{}(0, I_d)$. Let $G_{-i} = \frac{1}{n} \sum_{i=1}^n Z_i g'(Z_i^\top L^\top \beta) Z_i$. Noting that $\| L^{\top} \beta \|_2^2 = \beta^\top LL^\top \beta = \beta^\top \Sigma \beta = \gamma^2$, applying Lemma 7 of \cite{SurChCa19} gives
that $P\left( \lambda_{\min}( G_{-i} ) > \lambda_{lb} \right) \ge 1 - C\exp(-cn)$, for some $\lambda_{lb} > 0$.

Noting that the bounded condition number of $\Sigma$ implies that the operator norm of $L^{-\top}$ is bounded, we have that with probability converging to $1$,
\begin{equation*}
    \|H_{-i}^{-1} V_n\|_2^2 = \|\frac{1}{n} L^{-\top} G_{-i}^{-1} L^{-1} V_n\|_2^2 \le \frac{2 C^2 \kappa}{n^2 \lambda_{lb}^2}.
\end{equation*}
All of the above results hold with exponentially high probability, such that we can union bound over the $n$ remainder terms, for each $i$ and still have the probability converge to $1$.

Plugging all of these high probability bounds into the RHS of \eqref{eq:cs-bound} gives
\begin{equation*}
    \frac{1}{n} \sum_{i=1}^n \|X_i\|_2^2 n^{2\varepsilon}\|H_{-i}^{-1} V_n\|_2^2 \le 4p\frac{2 C^2 \kappa n^{2\varepsilon}}{n^2 \lambda_{lb}^2} = \frac{8 C^2 \kappa^2 n^{2\varepsilon}}{n \lambda_{lb}^2}
\end{equation*}
with probability converging to $1$. Similar derivation shows that
\begin{equation*}
    \left| \left(\frac{1}{n}\sum_{i=1}^n X_i^\top \what{\beta}_{-i}\right)^2 - \left(\frac{1}{n}\sum_{i=1}^n  X_i^\top \wt{\beta}_{-i}\right)^2 \right| \le 2 \kappa C \frac{n^{\varepsilon}}{\sqrt{n}},
\end{equation*}
also with probability going to $1$, and so $\what{\eta}_{\SLOE{}} = \what{\eta}^2_{\text{LOO}} + o_P(1)$.
\end{proof-of-proposition}

\section{Proofs of Lemmas}
\label{sec:lemma-proofs}
\begin{proof-of-lemma}[\ref{lem:loo-norm-diff}]
We use the following result from Lemma 18 from \citet{SurChCa19}. There, they define
\begin{gather*}
    q_i = \frac{1}{n} X_i^\top H_{-i}^{-1} X_i,\\
    \what{b} = \what{\beta}_{-i} - \frac{1}{n} H_{-i}^{-1} X_i\left(g(\prox_{q_i G}(X_i^\top \what{\beta}_{-i})) \right),
\end{gather*}
and show that
\begin{equation*}
    P\left( \| \what{\beta} - \what{b} \|_2 \le C \frac{K_n^2 H_n}{n} \right) \ge 1 - C n \exp( -c H_n^2) - C \exp(-c K_n^2) - \exp(-c n(1 + o(1))).
\end{equation*}

Additionally, in the proof (Eq. (165) and (172), respectively), they show that
\begin{equation*}
    P\left(\| H_{-i}^{-1} X_i \|_2^2 \le Cn\right) \ge 1 - \exp(-\Omega(n)).
\end{equation*}
and
\begin{equation*}
    P\left(g(\prox_{q_i G}(X_i^\top \what{\beta}_{-i})) \le C K_n \right) \ge 1 - C \exp(-C_3 K_n^2) - C \exp(-c n).
\end{equation*}
Together, these show that
\begin{align*}
    P\left( \|\what{b} - \what{\beta}_{-i}\|_2 \ge \frac{C^2 K_n}{\sqrt{n}} \right) &= 
    P\left(\frac{1}{n}\| H_{-i}^{-1} X_i g(\prox_{q_i G}(X_i^\top \what{\beta}_{-i})) \|_2 \le C K_n \frac{C}{\sqrt{n}} \right) \\
    &\ge 1 - \exp(-\Omega(n)) - C \exp(-c K_n^2) - C \exp(-c n).
\end{align*}

With this in mind, observe that
\begin{align*}
    \left\| \what{\beta} - \what{\beta}_{-i} \right\|_2 &= \left\| \what{\beta} - \what{b} + \what{b} - \what{\beta}_{-i} \right\|_2
    \\
    &\le \left\| \what{\beta} - \what{b} \right\|_2 + \left\| \what{b} - \what{\beta}_{-i} \right\|_2
    \\
    &\le C_1 \frac{K_n^2 H_n}{n} + \frac{C^2 K_n}{\sqrt{n}}
\end{align*}
with probability at least
\begin{equation*}
    1 - \exp(-\Omega(n)) - C \exp(-c K_n^2) - C n \exp( -c H_n^2) - C \exp(-c n(1 + o(1))),
\end{equation*}
for some $C, c > 0$, as claimed.
\end{proof-of-lemma}

\begin{proof-of-lemma}[\ref{lem:sup-loo}]
To prove the first statement, we will simply take a union bound over the $n(n-1)$ events for each pair of $i,k$ that
\begin{equation*}
    E_{ik} = \left\{ |X_i^\top (\what{\beta}_{-ik} - \what{\beta}_{-i})| \ge \frac{C K_n^2 H_n}{\sqrt{n}}\right\}.
\end{equation*}
Lemma 11 from the Supplementary Materials of \citet{SurCa2019} essentially shows that $P(E_{ik}) = o(1)$. We reproduce their Lemma 11 here for completeness. In this context, they assume that the $j$-th predictor is null, $\beta_j = 0$.

\begin{lemma}[Lemma 11, \citet{SurCa2019}]
For any pair $(i, k) \in [n]$, let $\what{\beta}_{-i, -j}, \what{\beta}_{-k,-j}$ denote the MLEs obtained on dropping the $i$-th and $k$-th observations respectively, and, in addition, removing the $j$-th predictor. Further, denore $\what{\beta}_{-ik,-j}$ to be the MLE obtained on dropping both the $i$-th, $k$-th observations and the $j$-th predictor. Then the following relation holds
\begin{equation*}
    P\left( \max\left\{ \left| X_{i,-j}^\top \left(\what{\beta}_{-i,-j} - \what{\beta}_{-ik, -j}\right) \right|, \left| X_{k,-j}^\top \left(\what{\beta}_{-k,-j} - \what{\beta}_{-ik, -j}\right) \right| \right\} \lesssim
n^{-1/2 + o(1)} \right) = 1 - o(1).
\end{equation*}
\end{lemma}
While they do not precisely track the rates of the lower order terms on the event or it's probability, inspecting their proof, which uses a slight modification of Lemma 17 and 18 from \citet{SurChCa19}, shows that the following precise bound holds:
Let $K_n$ and $H_n$ satisfy the conditions in \eqref{eq:sequence-cond}. Then, there exists universal constants $C_1, C_2, C_3, C_4, c_2, c_3 > 0$ such that
\begin{gather*}
    P\left( \max\left\{ \left| X_{i,-j}^\top \left(\what{\beta}_{-i,-j} - \what{\beta}_{-ik, -j}\right) \right|, \left| X_{k,-j}^\top \left(\what{\beta}_{-k,-j} - \what{\beta}_{-ik, -j}\right) \right| \right\} \le
 \frac{C_1 K_n^2 H_n}{\sqrt{n}} \right) \\
 \quad\,\quad\,\quad\,\quad\, \ge 1 - C_2 n \exp(-c_2 H_n^2) - C_3 \exp(-c_3 K_n^2) - \exp(-C_4 n(1 + o(1))).
\end{gather*}
A null predictor left out of fitting the MLE has no effect on the problem, so we can ignore the dependence on $j$, to get $P(E_{ik}) \le C n \exp(-c H_n^2) + C \exp(-c K_n^2) + \exp(-c n(1 + o(1))).$

Taking a union bound over the $n(n-1)$ events $E_{ik}$ proves the result of Lemma~\ref{lem:sup-loo}, which will be $o(1)$ under the conditions on $K_n$ and $H_n$ that $n^4 \exp(-c_1 H_n^2) = o(1)$, and $n^3 \exp(-c_2 K_n^2) = o(1)$, for any $c_1, c_2 > 0$ made in condition~\eqref{eq:sequence-cond}.

To prove the second statement, note that \citet[Theorem 4]{SurChCa19} implies that $\|\what{\beta}\|_2 > C$ with probability less than $ C \exp(-c n)$. Lemma~\ref{lem:loo-norm-diff} shows that $\what{\beta}_i$ is in a $K_n/\sqrt{n}$-neighborhood of $\what{\beta}$ with high probability. Together, these imply that $\|\what\beta_{-i}\|_2 > C$ with probability at most 
\begin{equation*}
  \exp(-\Omega(n)) + C \exp(-c K_n^2) + C n \exp(-c H_n^2) + C \exp(-cn(1 + o(1))).
\end{equation*}
Then, using that $X_i$ is independent of $\what{\beta}_{-i}$, we have
\begin{align*}
    P(|X_i^\top \what\beta_{-i}| > C^2 K_n ) &\le P( |X_i^\top \what{\beta}_{-i}| > C^2 \mid \|\what{\beta}_{-i}\|_2 \le C) + P(\|\what{\beta}_{-i}\|_2 > C)
    \\
    &= \E\left[ P(|X_i^\top \what{\beta}_{-i}| > C^2 K_n \mid \what{\beta}_{-i}) \mid \|\what{\beta}_{-i}\|_2 \le C\right] + P(\|\what{\beta}_{-i}\|_2 > C)
    \\
    &= \E\left[ P(|X_i^\top \what{\beta}_{-i}| > C^2 K_n \mid \what{\beta}_{-i}) \mid \|\what{\beta}_{-i}\|_2 \le C\right] + P(\|\what{\beta}_{-i}\|_2 > C)
    \\
    &\le \E\left[ C \exp(-c K_n^2) \mid \|\what{\beta}_{-i}\|_2 \le C\right] + P(\|\what{\beta}_{-i}\|_2 > C)
    \\
    &= C \exp(-c K_n^2) + P(\|\what{\beta}_{-i}\|_2 > C)
\end{align*}
where the last inequality follows from the fact that conditional on $\what{\beta}_{-i},$ $X_i^\top \what{\beta}_{-i} \sim \normal{}(0, \|\what{\beta}_{-i}\|_2^2)$, and uses the standard tail bound of a Gaussian distribution. Taking a union bound over $i \in \{1, \dots, n\}$ gives that complement of the statement in the Lemma occurs with probability less than
\begin{equation*}
    n(C \exp(-c K_n^2) + \exp(-\Omega(n)) + C \exp(-c K_n^2) + C n \exp(-c H_n^2) + C \exp(-cn(1 + o(1)))).
\end{equation*}
\end{proof-of-lemma}

\begin{proof-of-lemma}[\ref{lem:expectation-nearly-zero}]
We know that $\E[Z_k] = 0$, however, to control the expectation in \eqref{eq:expectation-nearly-zero}, we need to deal with two issues.  The first is that we need to ensure that all of the quantities in the expectation are absolutely integrable, so that all expectations are well-defined and finite. Then, the second is that the events $E_n$ and $B_n$ on which $\goodE[\cdot] = \E[\cdot \mid E_n \cap B_n]$ is conditioned are not independent of $Z_k$, so we must check that conditioning does not change the expectation too much.

To address the first issue, we will condition on the event that the leave-one-out MLE $\what{\beta}_{-k}$ and all of the leave-two-out MLEs leaving out $k$, $\{\what{\beta}_{-ik}\}_{i\not= k}$, are bounded:
\begin{equation*}
    V_k = \{ \|\what{\beta}_{-k}\|_2 \le C, \sup_{i \not= k} \|\what{\beta}_{-ik}\|_2 \le C\},
\end{equation*}
where $C$ is chosen so that $V_k \supset B_n$ (that is, $B_n$ implies $V_k$) for all $n,k$.
Notice that $V_k$ is independent of $Z_k$, and ensures that all of the quantities in \eqref{eq:expectation-nearly-zero} are sufficiently bounded or integrable, so that
\begin{equation*}
    \E[ \what{\beta}_{-ik}^\top Z_i \what{\beta}_{-ik} \what{\beta}_{-k}^\top Z_k \what{\beta}_{-k} \mid V_k ] = 0.
\end{equation*}

Now, we relate $\goodE[ \what{\beta}_{-ik}^\top Z_i \what{\beta}_{-ik} \what{\beta}_{-k}^\top Z_k \what{\beta}_{-k}]$ to $\E[ \what{\beta}_{-ik}^\top Z_i \what{\beta}_{-ik} \what{\beta}_{-k}^\top Z_k \what{\beta}_{-k} \mid V_k ]$ by splitting up the latter into parts conditional on $E_n \cup B_n$ and $(E_n \cap B_n)^C$. Indeed,
\begin{align*}
  \E[ \what{\beta}_{-ik}^\top Z_i \what{\beta}_{-ik} \what{\beta}_{-k}^\top Z_k \what{\beta}_{-k} \mid V_k] &= \E[ \what{\beta}_{-ik}^\top Z_i \what{\beta}_{-ik} \what{\beta}_{-k}^\top Z_k \what{\beta}_{-k} \ind{E_n \cap B_n} \mid V_k ] \\
  &~~~~~~~~~ + \E[ \what{\beta}_{-ik}^\top Z_i \what{\beta}_{-ik} \what{\beta}_{-k}^\top Z_k \what{\beta}_{-k} \ind{(E_n \cap B_n)^C} \mid V_k ],
\end{align*}
and recalling that $V_k \supset (E_n \cap B_n)$, we know that
\begin{equation*}
\E[ \what{\beta}_{-ik}^\top Z_i \what{\beta}_{-ik} \what{\beta}_{-k}^\top Z_k \what{\beta}_{-k} \ind{E_n \cap B_n} \mid V_k ] = \E[ \what{\beta}_{-ik}^\top Z_i \what{\beta}_{-ik} \what{\beta}_{-k}^\top Z_k \what{\beta}_{-k} \mid E_n \cap B_n ] P(E_n \cap B_n \mid V_k).
\end{equation*}

Using that $\E[ \what{\beta}_{-ik}^\top Z_i \what{\beta}_{-ik} \what{\beta}_{-k}^\top Z_k \what{\beta}_{-k} \mid V_k ]  = 0$,
\begin{align*}
    \left| \E[ \what{\beta}_{-ik}^\top Z_i \what{\beta}_{-ik} \what{\beta}_{-k}^\top Z_k \what{\beta}_{-k} \mid E_n \cap B_n ] P(E_n \cap B_n \mid V_k) \right| &= \left| \E[ \what{\beta}_{-ik}^\top Z_i \what{\beta}_{-ik} \what{\beta}_{-k}^\top Z_k \what{\beta}_{-k} \ind{(E_n \cap B_n)^C} \mid V_k ] \right|
    \\
    \left| \E[ \what{\beta}_{-ik}^\top Z_i \what{\beta}_{-ik} \what{\beta}_{-k}^\top Z_k \what{\beta}_{-k} \mid E_n \cap B_n ] \right| &= \frac{\left| \E[ \what{\beta}_{-ik}^\top Z_i \what{\beta}_{-ik} \what{\beta}_{-k}^\top Z_k \what{\beta}_{-k} \ind{(E_n \cap B_n)^C} \mid V_k ] \right|}{P(E_n \cap B_n \mid V_k)}
\end{align*}
Now, in the proof of Theorem~\ref{thm:consistency}, we showed that Lemma~\ref{lem:sup-loo} and Lemma~\ref{lem:loo-norm-diff} imply that $P(E_n \cap B_n) \to 1$, and because $V_k \supset (E_n \cap B_n)$, this implies that $P(E_n \cap B_n \mid V_k) \to 1$, as well. What remains is to control the numerator of the previous display.


Applying the Cauchy-Schwarz inequality gives
\begin{equation*}
\left| \E[ \what{\beta}_{-ik}^\top Z_i \what{\beta}_{-ik} \what{\beta}_{-k}^\top Z_k \what{\beta}_{-k} \ind{(E_n \cap B_n)^C} \mid V_k ] \right| \le \sqrt{\E\left[ \left(\what{\beta}_{-ik}^\top Z_i \what{\beta}_{-ik} \what{\beta}_{-k}^\top Z_k \what{\beta}_{-k}\right)^2 \mid V_k \right] P\left((E_n \cap B_n)^C \mid V_k\right)}.
\end{equation*}
A very loose bound on the first term, using that $\|X_i\|_2^2 \lesssim p$ with high probability, $\what{\beta}_{-ik} \le C$ on $V_k$, and similar expressions hold for the terms involving $\what{\beta}_{-k}$ and $X_k$ gives
\begin{equation*}
    \E\left[ \left(\what{\beta}_{-ik}^\top Z_i \what{\beta}_{-ik} \what{\beta}_{-k}^\top Z_k \what{\beta}_{-k}\right)^2 \mid V_k \right] \lesssim p^4. 
\end{equation*}
Because $V_k \supset (E_n \cap B_n)$, we know that $P((E_n \cap B_n)^C \mid V_k) \le P((E_n \cap B_n)^C) \lesssim n^2 \exp(-c_2 H_n^2) + n\exp(-c_3 K_n^2) + n \exp(-C_4n(1 + o(1))) + \exp(-\Sigma(n))$.
Altogether, we have 
\begin{align*}
    &n \left| \E[ \what{\beta}_{-ik}^\top Z_i \what{\beta}_{-ik} \what{\beta}_{-k}^\top Z_k \what{\beta}_{-k} \ind{(E_n \cap B_n)^C} \mid V_k ] \right|
    \\ &~~~~~ \lesssim \sqrt{ n^2 p^4 \left(n^2 \exp(-c_2 H_n^2) + n\exp(-c_3 K_n^2) + n \exp(-C_4n(1 + o(1))) + \exp(-\Sigma(n))\right)}.
\end{align*}
Using the conditions on $K_n$ and $H_n$ from \eqref{eq:sequence-cond}, we know that this bound goes to $0$, completing the proof.%
\end{proof-of-lemma}

\section{Genomics}
Variants known to be associated with glaucoma, in the form ``(chromosome)-(position)-(allele1)-(allele2)'' using coordinates from the GRCh37 human genome build. 111 in total.
\begin{center}
\begin{longtable}{ |c|c|c|c|c| } 
\hline
01-101095202-A-G & 01-113242122-T-G & 01-162679145-G-A & 01-165737704-C-G & 01-171605478-G-A \\ 
01-219215137-G-A & 01-36612955-C-A & 01-38076621-C-T & 01-54123873-G-T & 01-68837169-A-C \\ 
01-8495590-A-G & 01-88213014-T-C & 02-111638775-C-T & 02-12951321-C-T & 02-153364527-A-G \\ 
02-213760746-AT-A & 02-28365914-G-A & 02-45878760-G-T & 02-55933014-C-T & 02-59523041-T-C \\ 
02-66537344-G-T & 02-69411517-A-C & 02-71651939-T-A & 03-105073472-A-C & 03-169239578-A-G \\ 
03-171821356-A-G & 03-186128816-A-G & 03-188066953-T-G & 03-24510794-A-C & 03-25581798-C-T \\ 
03-56876596-T-C & 03-85172364-G-C & 04-184779187-G-T & 04-54027595-A-G & 04-7904363-G-A \\ 
05-55783678-G-A & 06-122645298-A-C & 06-134372150-C-G & 06-136462744-T-G & 06-1548369-A-G \\ 
06-158971266-A-G & 06-170454915-A-G & 06-36570366-T-C & 06-51414922-C-T & 07-103624813-A-G \\ 
07-116162306-A-T & 07-11679113-A-G & 07-117636111-C-G & 07-134520521-C-A & 07-151505698-C-T \\ 
07-28401455-C-G & 07-35961137-C-T & 07-39077397-C-T & 07-80845529-G-GA & 07-82949529-T-G \\ 
08-108273318-T-G & 08-124554317-A-T & 08-30454209-CA-C & 08-6377141-C-G & 09-107695539-T-C \\ 
09-129390800-C-T & 09-22051670-G-C & 10-10840849-A-C & 10-115546535-A-G & 10-126278648-T-C \\ 
10-60326910-G-A & 10-78282063-T-C & 10-94929116-C-T & 10-96023077-T-C & 11-102064834-C-A \\ 
11-115039683-G-A & 11-128380742-C-A & 11-130282078-T-C & 11-17011176-C-A & 11-47469439-A-G \\ 
11-65337251-A-T & 11-86368106-T-C & 12-107219308-A-G & 12-111932800-C-T & 12-28203245-T-A \\ 
12-83948055-T-C & 13-110777939-C-G & 13-22673870-A-G & 13-76258720-A-G & 14-53960089-A-G \\ 
14-75084829-G-A & 14-76371658-G-C & 14-95956875-T-C & 15-57553832-A-T & 15-61947280-C-G \\ 
15-67025403-C-T & 15-74221298-C-T & 15-92331707-A-G & 16-51601948-C-T & 16-59995564-A-G \\ 
16-65067443-C-T & 16-77661732-C-T & 17-10031183-A-G & 17-2201944-A-G & 17-44025888-C-A \\ 
20-38074218-T-C & 20-45534053-A-G & 20-6470094-G-A & 21-27216839-T-A & 21-40406630-G-A \\ 
22-19870147-C-T & 22-29108229-A-G & 22-38176979-T-G & X-13954397-C-T & X-3329593-C-T \\ 
X-43940827-T-C & & & & \\ 
\hline
\end{longtable}
\end{center}

\end{document}